\documentclass[11pt,a4paper]{article}
\usepackage{times,latexsym}
\usepackage{url}
\usepackage[T1]{fontenc}
\usepackage{graphicx}
\usepackage{amsmath}
\usepackage{amssymb}
\usepackage{booktabs}
\usepackage{multirow}
\usepackage{colortbl}
\usepackage{xcolor}
\usepackage[most]{tcolorbox}
\tcbuselibrary{breakable}
\usepackage{float}
\usepackage{changepage}

\usepackage[acceptedWithA]{tacl2021v1}

\usepackage{xspace,mfirstuc,tabulary}

\newif\iftaclinstructions
\taclinstructionsfalse 
\iftaclinstructions
\renewcommand{\confidential}{}
\renewcommand{\anonsubtext}{(No author info supplied here, for consistency with
TACL-submission anonymization requirements)}
\newcommand{\instr}
\fi

\iftaclpubformat 

\else

\fi

\title{ACE-RL: Adaptive Constraint-Enhanced Reward for Long-form Generation Reinforcement Learning}

\author{
    Jianghao Chen\textsuperscript{1,2,3} \
    Wei Sun\textsuperscript{1,2} \
    Qixiang Yin\textsuperscript{3} \
    Zhixing Tan\textsuperscript{4} \
    Jiajun Zhang\textsuperscript{1,2,5}\thanks{\ \ Corresponding Author} \\
    \textsuperscript{1}Institute of Automation, Chinese Academy of Sciences \\
    \textsuperscript{2}School of Artificial Intelligence, University of Chinese Academy of Sciences \\
    \textsuperscript{3}Zhongguancun Academy, Beijing, China \textsuperscript{4}Tsinghua University \textsuperscript{5}Wuhan AI Research \\
    \{chenjianghao2022, sunwei2023\}@ia.ac.cn, s-yqx24@bjzgca.edu.cn,\\
    tzx.2019@tsinghua.org.cn, jjzhang@nlpr.ia.ac.cn
}

\date{}

\begin{document}
\maketitle
\begin{abstract}
Long-form generation has become a critical and challenging application for Large Language Models (LLMs). Existing studies are limited by their reliance on scarce, high-quality long-form response data and their focus on coarse-grained, general-purpose metrics (e.g., coherence and helpfulness), overlooking the nuanced, scenario-specific requirements of real-world tasks. To address these limitations, we propose a framework utilizing \textbf{A}daptive \textbf{C}onstraint-\textbf{E}nhanced reward for long-form generation \textbf{R}einforcement \textbf{L}earning (ACE-RL). ACE-RL first decomposes each instruction into a set of fine-grained, adaptive constraint criteria spanning key dimensions of long-form generation tasks. Subsequently, we design a reward mechanism to quantify the response quality based on their satisfaction over corresponding constraints, converting subjective quality evaluation into constraint verification. Finally, we leverage reinforcement learning to optimize LLMs using these fine-grained signals. Experimental results show that ACE-RL significantly outperforms existing SFT and RL baselines by 18.63\% and 7.61\% on WritingBench, and our top-performing model even surpasses proprietary systems like GPT-4o by 8.76\%, providing a more effective training paradigm in long-form generation scenarios.
\end{abstract}

\section{Introduction}
Recent advancements in Large Language Models (LLMs) have yielded significant improvements in their long-context understanding capabilities \citep{comanici2025gemini25pushingfrontier, GPT-4o, claude-4, yang2025qwen3}. This success has naturally pushed the research frontier towards a more challenging field: high-quality long-form generation (e.g., storytelling, report writing, or legal document drafting). However, generating coherent, well-structured, and engaging content over thousands of words still remains a significant and unresolved challenge for current long-context LLMs \citep{wu2025shifting, wu2025longgenbench, bai2025longwriter}.

\begin{figure}[t]
  \centering
  \includegraphics[width=0.9\columnwidth]{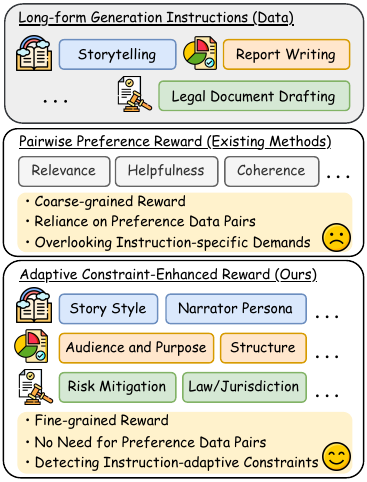}
  \caption{Comparison of reward mechanisms: conventional methods vs. our proposed method.}
  \label{motivation}
  \vspace{-2mm}
\end{figure}

While many researchers focus on supervised fine-tuning (SFT) \citep{bai2025longwriter, pham2024suri}, these approaches are often constrained by reliance on costly synthetic datasets from proprietary LLMs and a lack of optimization for response quality attributes \citep{deng2022model, pham2024suri}. In contrast, reinforcement learning (RL) alleviates this by enabling direct optimization of desired qualities through a reward mechanism. This allows for a more flexible and robust approach to training models for long-form generation, particularly when focusing on specific human preferences. However, existing RL-based methods primarily rely on pairwise preference reward with coarse-grained evaluation dimensions \citep{lei2025writing, wu2025longwriter}. As shown in Figure \ref{motivation}, while these methods ensure fundamental qualities such as relevance, coherence, and helpfulness, they fall short of addressing the detailed, instruction-adaptive demands of diverse tasks. For example, given a storytelling instruction: "Write a 3000-word novel about a young, struggling couple with an O. Henry-style ending." The core of the instruction is the writing style of a surprise, ironic twist, a quality that is not a universal marker of "good" writing, but a highly specific stylistic demand. Moreover, the reliance on extra preference response pairs imposes a significant data collection burden, making the process both costly and difficult to scale.

To address the above issue, we propose a framework with \textbf{A}daptive \textbf{C}onstraint-\textbf{E}nhanced reward for long-form generation \textbf{R}einforcement \textbf{L}earning (ACE-RL). The core innovation of ACE-RL lies in its training paradigm shift away from coarse-grained, subjective rewards toward fine-grained, verifiable rewards. Specifically, ACE-RL begins by employing an automated pipeline to collect and analyze diverse long-form generation instructions from real-world human-LLM interactions. This pipeline deconstructs each instruction by identifying both its explicit demands and implicit intents, converting them into a checklist of verifiable constraints. These constraints capture multifaceted dimensions, including content completeness, structural logic, and stylistic formatting. We then introduce a novel reward mechanism that scores a generated response based on its satisfaction with these constraints. This process effectively transforms the subjective challenge of evaluating writing quality into a series of constraint verification tasks. Finally, by leveraging reinforcement learning guided by this fine-grained reward signal, ACE-RL significantly enhances the long-form generation capability of LLMs.

Our contributions are summarized as follows:
\begin{itemize}
\item We propose ACE-RL, a novel reinforcement learning framework that utilizes fine-grained, instruction-adaptive rewards for long-form generation.
\item We introduce an automatic pipeline for constructing long-form generation instructions with constraint checklists, and we will release our datasets for future studies. 
\item We design a novel reward mechanism that transforms subjective quality evaluation of long-form responses into constraint verification tasks.
\item Extensive experimental results demonstrate that our method significantly outperforms SFT and RL baselines, and even surpasses strong proprietary LLMs.
\end{itemize}

\begin{figure*}[t]
  \includegraphics[width=\linewidth]{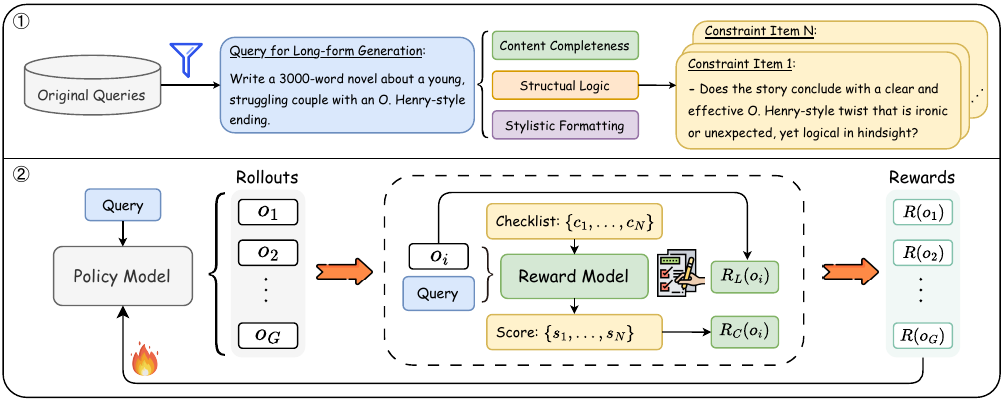}
  \caption{The overall framework of ACE-RL. First, we collect diverse instructions for long-form generation tasks and create an instruction-adaptive constraint checklist for each across three dimensions. Second, a reward model is deployed to verify whether the policy model's responses meet each constraint. This constraint-enhanced reward, along with a length reward, are then used for RL training.}
  \label{framework}
\end{figure*}

\section{Related Work}
\subsection{Long-context Understanding for LLMs}
The enhanced long-context modeling capabilities of LLMs have significantly broadened their application across various fields \citep{liu2025comprehensive, jin2025longcontext}. Recent studies have proposed both model-centric and data-centric methods. Model-centric methods involve novel model architectures \citep{gu2024mamba, dao2024transformers, peng2023rwkv}, efficient attention \citep{liu2024deepseek, lu2025moba, shah2024flashattention}, and extrapolation of positional encoding \citep{xiong2024effective, peng2024yarn, ding2024longrope, an2024trainingfree}. These approaches focus on redesigning the core components of LLMs to handle extensive contexts more efficiently and effectively. In parallel, data-centric methods improve the long-context ability of LLMs through continual pre-training \citep{fu2024data, chen2024long, chen2025ladm} or supervised fine-tuning \citep{bai2024longalign, an2024make, chen2024longlora}. However, these studies mainly focus on long-context input scenarios, and generating high-quality long-form outputs remains a significant challenge for current LLMs.

\subsection{Long-form Generation for LLMs}
The challenge of long-form generation has been approached primarily from two directions: supervised fine-tuning and reinforcement learning. SFT-based methods focus on data synthesis, such as long-form response construction \citep{quan2024language, bai2025longwriter} and instruction back-translation \citep{pham2024suri}. Recent success of reinforcement learning with preference reward \citep{sun2025rethinking, yang-etal-2025-implicit} encourages researchers to explore outcome-based RL for long-form generation. \citet{lei2025writing} leverage proprietary LLMs to generate reference responses and calculate pairwise reward by quality comparison of model responses and reference responses. \citet{wu2025longwriter} utilize a reward model trained with preference data pairs to capture general response quality such as coherence and helpfulness. 

However, these approaches rely on coarse-grained rewards with extra preference data pairs, failing to capture the fine-grained, instruction-specific details that define high-quality writing in diverse scenarios. Therefore, we propose the ACE-RL framework, which utilizes a novel reward mechanism with fine-grained, instruction-adaptive constraint verification.

\section{Methodology}
\label{method}
This section details the framework of our proposed ACE-RL, a novel approach for improving the long-form generation capabilities of LLMs. As illustrated in Figure \ref{framework}, the core innovation lies in systematically breaking down each high-level user instruction into a checklist of fine-grained constraints. This process transforms the abstract goal of "writing a good response" into a set of concrete, verifiable objectives, enabling us to construct a precise reward signal that directly reflects the model's ability to meet specific user demands.

\subsection{Data Preparation}
\label{preparation}
To construct a high-quality dataset tailored for long-form generation, we collect a diverse set of instructions from WildChat-1M \citep{zhao2024wildchat}, a large-scale dataset of real-world human-LLMs interactions. We first conduct deduplication to remove redundant instructions. Then, we employ Qwen3-235B-A22B \citep{yang2025qwen3} to filter out instructions that specifically call for long-form generation. Furthermore, considering that adherence to length constraints is a critical aspect of long-form generation tasks, we augment each instruction with the target word count by leveraging the same LLM for analysis. Detailed prompts are shown in Appendix \ref{prompt}.

\begin{figure*}[!h]
  \includegraphics[width=\linewidth]{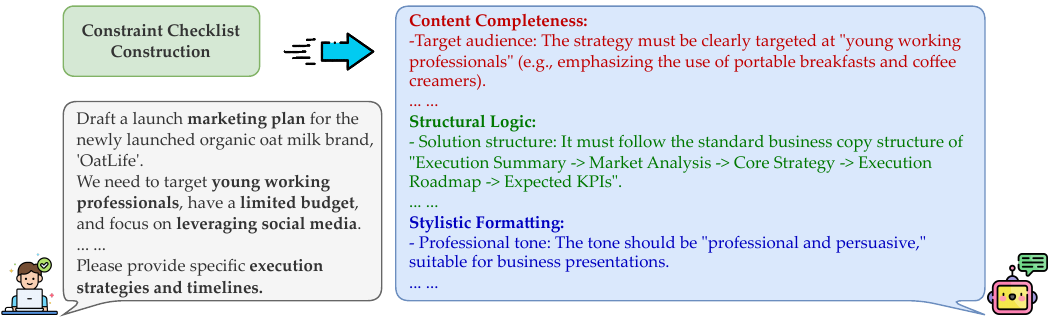}
  \caption{Examples of constraint generation from real-world user instructions across three key dimensions.}
  \label{example}
  \vspace{-2mm}
\end{figure*}

\subsection{Adaptive Constraint Construction}
\label{constraint}
Long-form generation instructions are often open-ended and under-specified, requiring models to not only address explicit requests but also to infer and fulfill a wide range of implicit expectations related to logical flow, depth of analysis, and completeness. Therefore, to facilitate fine-grained, adaptive evaluation of such tasks, ACE-RL employs a pipeline that automatically deconstructs a high-level user instruction into a checklist of verifiable constraints. This procedure transforms the ambiguous goal of a long-form generation instruction into a set of concrete, measurable objectives that can directly guide the RL process. We mainly focus on the following three dimensions:

\paragraph{Content Completeness} This dimension focuses on ensuring the generated output thoroughly addresses all components of the user's instruction. Constraints are generated to verify the inclusion of all explicitly requested topics, entities, key arguments, or questions posed in the prompt. Beyond explicit requests, this dimension also accounts for implicit expectations regarding the depth and breadth of information. These constraints measure whether the model has provided a comprehensive and factually grounded response that directly fulfills the core informational goals of the task.

\paragraph{Structural Logic} Long-form generation typically requires global structural organization and coherent logical progression across paragraphs. To capture this dimension, we generate constraints that describe the expected macro-structure, like introduction–body–conclusion or problem–analysis–solution, and the necessary coverage of key subtopics. This dimension explicitly targets common failure modes in long-form generation, such as drifting off-topic, repeating content, or failing to reach a conclusion.

\paragraph{Stylistic Formatting} Different long-form tasks demand distinct stylistic and formatting properties. The constraint checklist thus includes items for tone (e.g., formal, conversational, narrative), narrative perspective, and domain-specific style (e.g., legal, academic, journalistic). It also specifies formatting requirements such as the presence of section headings, bullet lists, citations, or code blocks when applicable. 




By systematically generating this constraint checklist for each instruction, we create a fine-grained blueprint for evaluating the quality of a long-form response (see Figure \ref{example} for an illustration). This serves as the foundation for our reward mechanism. We leverage Qwen3-235B-A22B for analysis and constraint generation. The detailed prompt and cases of instruction-checklist pairs are provided in Appendix \ref{prompt} and Appendix \ref{checklist_example}, respectively. To validate quality, we conducted a human evaluation on 30 randomly selected samples. The resulting scores of 82\% for coverage and 89\% for precision confirm the effectiveness of our constraint generation method. Detailed information is provided in Appendix \ref{checklist_quality}. However, we observed that examples with simple constraints were ineffective for RL training, as they yielded nearly identical high scores across all rollout responses. To mitigate this, we sample eight responses from Qwen2.5-7B \citep{yang2024qwen2} for each instruction and use Qwen3-235B-A22B for constraint verification (see Section \ref{reward}). We then filter out instructions for which the average verification score across the eight responses exceeds 0.85. We provide a detailed analysis in Section \ref{data_quality}.



\subsection{Reward Design and RL Training}
\label{reward}
The design of the reward function is crucial for guiding policy model optimization during RL training. Recent studies often employ LLM-as-a-Judge for evaluating the quality of long-form outputs \citep{lei2025writing, wu2025longwriter}. These approaches typically use abstract, holistic quality evaluation derived from pairwise preference comparisons. Such methods remain subjective, coarse-grained, and fail to capture the fine-grained, specific demands of user instructions. In contrast, we utilize the satisfaction of model responses over the constraint checklists as the reward signals, providing a more fine-grained and concrete guidance for the policy optimization process. We explain the details of the algorithm and reward design for RL training as follows.

\paragraph{Training Algorithm}  We adopt the Group Relative Policy Optimization (GRPO) \citep{shao2024deepseekmath} algorithm for RL training. GRPO removes the additional value model from PPO \citep{schulman2017proximal} and instead calculates the advantages through group comparison. Given a specific query $q$ from training dataset $\mathcal{D}$, GRPO samples $G$ rollouts $\{o_1, \dots, o_G\}$ from the old policy model $\pi_{\theta_{\mathrm{old}}}$ and calculate their rewards $\{R_1, \dots, R_G\}$. Then the advantage $A_i$ of each rollout is estimated by comparing each $R_i$ with the average reward of this group. The final optimization objective of GRPO is defined as follows:

\begin{equation}
\small
\begin{aligned}
r_i = \frac{\pi_\theta(o_i \mid q)}{\pi_{\theta_{\mathrm{old}}}(o_i \mid q)}, \, r_i^{\mathrm{c}} = \mathrm{clip}(r_i, 1-\varepsilon, 1+\varepsilon),
\end{aligned}
\label{eq:r}
\end{equation}

\begin{equation}
\small
\begin{aligned}
A_i = \frac{R_i - \operatorname{mean}(\{R_1, \dots, R_G\})}{\operatorname{std}(\{R_1, \dots, R_G\})},
\end{aligned}
\label{eq:advantage}
\end{equation}

\begin{equation}
\small
\begin{aligned}
\mathcal{J}(\theta) 
 & = \mathbb{E}_{q \sim \mathcal{D}, \{o_i\}_{i=1}^G \sim \pi_{\theta_{\mathrm{old}}}(\cdot\mid q)} \\ 
 & \hspace*{-2.5em}\Biggl[\frac{1}{G} \sum_{i=1}^G \Bigl( \min( r_iA_i, r_i^{\mathrm{c}}A_i ) - \beta D_{\mathrm{KL}}(\pi_{\theta} \| \pi_{\mathrm{ref}}) \Bigr) \Biggr],
\end{aligned}
\label{eq:grpoloss}
\end{equation}
where $r_i$ is the importance sampling ratio, $\varepsilon$ is the clip threshold, $\beta$ controls the penalty strength of KL divergence, and $\pi_{\mathrm{ref}}$ is the fixed reference policy model.

\paragraph{Length Reward}
In long-form generation, meeting a specified length is often a key requirement. Existing LLMs tend to provide too short responses or fall into endless repetition \citep{bai2025longwriter}. Therefore, we introduce a length reward to control the generation length for the response of the policy model. Given an instruction with target length $L_t$ and a rollout response $\hat{y}$, we define the follwing length reward $R_L(\hat{y})$:
\begin{gather}
\delta = \frac{\lvert L_{\hat{y}} - L_t \rvert}{L_t} \label{delta}, \\
R_L(\hat{y}) = 
\begin{cases}
    1.0 & \delta \leq \Delta \\
    \exp\left(-\alpha \cdot \left( \delta - \Delta\right)\right) & \delta > \Delta
\end{cases} \label{length},
\end{gather}
where $\delta$ is the relative length deviation, $\Delta$ is the allowable relative deviation threshold, and the decay factor $\alpha$ controls the degree of length penalty. One rollout response will get a maximum reward of 1.0 if its length falls within an acceptable range of the target length. If the length deviates beyond this range, the reward would decrease exponentially.

\paragraph{Constraint Reward}
To evaluate whether one rollout response meets each item in the generated constraint checklist, we employ a verifier LLM. Specifically, given an instruction $Inst$, its corresponding constraint checklist $\{c_1, c_2, \dots, c_N\}$ and policy model response $\hat{y}$, the verifier model is prompt to output a three-level judgment: \texttt{"Fully Met"} if the response completely satisfies the constraint; \texttt{"Partially Met"} if the response addresses the constraint to some extent but is incomplete or contains minor errors; \texttt{"Not Met"} if the response fails to address the constraint. This transforms the subjective evaluation for long-form outputs into a series of constraint verification tasks, where utilizing the long-context understanding capability to guide the enhancement of the long-form generation ability of LLMs. We define the constraint reward $R_C(\hat{y})$ as following:
\begin{gather}
\hspace*{-0.5em}s(\hat{y}, c_i) = f_{\mathrm{LLM}}(T, Inst, \hat{y}, c_i) \in \{0, 0.5, 1\}, \label{s_i} \\
R_C(\hat{y}) = \frac{1}{N} \displaystyle \sum_{i=1}^{N} s(\hat{y}, c_i), \label{r_con}
\end{gather}
where $s(\hat{y}, c_i)$ is the score for one rollout response $\hat{y}$ to instruction $Inst$ on constraint $c_i$, which is determined by the verifier LLM $f_{\mathrm{LLM}}$ with prompt template $T$. The verifier LLM outputs a score from $\{0, 0.5, 1\}$, corresponding to the three-level judgments: \texttt{"Fully Met"}, \texttt{"Partially Met"}, and \texttt{"Not Met"}. The overall constraint reward is then calculated as the mean of all individual constraint scores. Through RL training, this reward signal guides the policy model to align its responses with the fine-grained, multifaceted demands of the user instructions. The prompt used for verifier LLM is shown in Appendix \ref{prompt}. 

To balance efficiency and accuracy during RL training, we utilize Qwen3-8B for constraint verification. To validate its reliability, we conducted a human evaluation on 200 randomly sampled instances, each comprising an instruction, a constraint, and a model response. Two annotators (PhD candidates in Computer Science) classified the responses as \texttt{"Fully Met"}, \texttt{"Partially Met"}, or \texttt{"Not Met"}, following the same definitions used in the verifier prompt. The model demonstrated agreement rates of 75.5\% and 77.0\% with the respective annotators, approaching the inter-human consistency of 84.5\%. Notably, the model exhibited minimal confusion between distinct categories, with an error rate of only 1.5\% when distinguishing between \texttt{"Fully Met"} and \texttt{"Not Met"} instances. Given that RL relies primarily on relative ranking rather than absolute scoring precision \citep{christiano2017deep}, these results confirm that Qwen3-8B serves as a sufficiently accurate proxy for constraint verification in our RL training pipeline.



\paragraph{Overall Reward} The overall reward $R(\hat{y})$ is defined as the average of the length reward and the constraint reward:
\begin{equation}
\begin{aligned}
R(\hat{y}) = \frac{1}{2}\Bigl(R_L(\hat{y}) + R_C(\hat{y})\Bigr).
\end{aligned}
\label{eq:overall_reward}
\end{equation}

\section{Experimental Settings}
\subsection{Training Dataset}
We conduct RL training on long-form generation instruction datasets constructed in Section \ref{method}. Each instruction is paired with a checklist of constraints and a length requirement. Table \ref{tab:dataset_statistics} shows the detailed statistics of our training dataset.

\begin{table}[h!]
\centering
\begin{tabular}{lr}
\toprule
\textbf{Statistic} & \textbf{Value} \\
\midrule
Total Number of Instructions & 18.1K \\
Average Number of Constraints & 5.18 \\
Average Required Length & 2005 \\
\bottomrule
\end{tabular}
\caption{Statistics of our training dataset for ACE-RL.}
\label{tab:dataset_statistics}
\end{table}

\subsection{Training Configuration} 
We perform RL training under the VERL \citep{sheng2025hybridflow} framework with the GRPO algorithm. We split 480 samples to construct a validation set and train each model for 200 steps, using a batch size of 64, a rollout number of 32, and a learning rate of $1 \times 10^{-6}$. To accommodate long-form generation, we set the maximum generation length to 8192 and use a sampling temperature of 1.0. For length reward, we set the relative deviation threshold $\Delta$ to 0.2 and the decay factor $\alpha$ to 0.5. For constraint verification, we utilize Qwen3-8B \cite{yang2025qwen3} for a balance between efficiency and performance. We also remove the KL penalty term following recent studies \citep{yu2025dapo}.

\begin{table*}[!ht]
\centering
\footnotesize
\resizebox{1.0\textwidth}{!}{
\begin{tabular}{l|c|cccccc|cc|cc|cc}
\toprule
\multirow{2.5}{*}{Models} & \multirow{2.5}{*}{\textbf{Avg}} & \multicolumn{6}{c|}{\textbf{Domains}} & \multicolumn{6}{c}{\textbf{Requirements}} \\
 \cmidrule(lr){3-8} \cmidrule(lr){9-14}
& & D1 & D2 & D3 & D4 & D5 & D6 & R1 & \multicolumn{1}{c|}{C} & R2 & \multicolumn{1}{c|}{C} & R3 & \multicolumn{1}{c}{C} \\ 
\midrule
\rowcolor{gray!15} \multicolumn{14}{l}{\textit{\textbf{Proprietary LLMs}}} \\
\midrule
o3-2025-04-16 & 85.27 & 84.81 & 85.20 & 83.89 & 85.88 & 85.82 & 86.80 & 85.14 & 87.45 & 85.24 & 90.98 & 86.31 & 87.21 \\
Gemini-2.5-pro-preview & 83.05 & 83.21 & 81.47 & 83.00 & 84.52 & 84.49 & 82.14 & 83.58 & 86.49 & 83.86 & 90.45 & 83.35 & 83.98 \\
Claude-3-7-sonnet & 78.48 & 78.24 & 77.93 & 76.51 & 79.37 & 79.26 & 80.88 & 79.43 & 82.51 & 78.84 & 86.14 & 79.23 & 80.49 \\
GPT-4o & 75.46 & 74.40 & 73.42 & 74.38 & 77.91 & 75.86 & 78.08 & 76.82 & 81.57 & 75.82 & 85.46 & 76.13 & 76.73 \\
o1-Preview & 68.57 & 68.54 & 67.01 & 66.57 & 69.53 & 70.31 & 71.41 & 70.09 & 75.10 & 68.49 & 79.78 & 70.91 & 73.81 \\

\midrule
\rowcolor{gray!15} \multicolumn{14}{l}{\textit{\textbf{Open-source LLMs}}} \\
\midrule
Deepseek-R1-0528 & 83.22 & 83.15 & 81.48 & 81.55 & 85.68 & 84.14 & 84.44 & 84.24 & 87.27 & 83.72 & 89.35 & 83.83 & 82.70 \\
Qwen3-235B-A22B-thinking & 81.45 & 80.19 & 79.24 & 80.95 & 82.92 & 82.52 & 82.89 & 82.54 & 85.02 & 81.30 & 88.22 & 81.26 & 81.76 \\
LongWriter-Zero-32B & 80.30 & 80.66 & 80.27 & 80.21 & 76.09 & 83.55 & 81.02 & 79.89 & 83.38 & 80.77 & 86.82 & 80.23 & 82.05 \\
Qwen3-235B-A22B & 73.63 & 73.56 & 72.90 & 73.98 & 70.13 & 76.52 & 74.69 & 77.46 & 82.05 & 77.01 & 87.29 & 76.26 & 79.57 \\
Qwen2.5-72B-instruct & 65.28 & 65.80 & 63.36 & 63.80 & 62.75 & 68.07 & 67.91 & 65.81 & 70.49 & 65.92 & 78.65 & 66.38 & 67.95 \\
LongWriter-glm-9B & 62.94 & 64.06 & 63.66 & 62.35 & 61.26 & 65.03 & 61.30 & 62.79 & 66.70 & 63.60 & 74.82 & 63.37 & 65.88 \\
LongWriter-llama3.1-8B & 58.01 & 60.05 & 59.31 & 57.58 & 56.03 & 58.38 & 56.73 & 58.12 & 61.40 & 58.60 & 67.61 & 59.05 & 62.97 \\
Llama-3.3-70b-instruct & 50.43 & 50.67 & 49.25 & 47.90 & 48.52 & 52.92 & 56.56 & 50.71 & 50.71 & 50.38 & 50.38 & 51.08 & 51.08 \\
\midrule
\midrule

Llama-3.1-8B-instruct & 47.10 & 48.67 & 45.91 & 44.60 & 43.53 & 48.86 & 51.02 & 46.66 & 50.29 & 46.75 & 55.71 & 46.48 & 44.82 \\
\quad\textit{w/ LongWriter SFT} & 58.01 & 60.05 & 59.31 & 57.58 & 56.03 & 58.38 & 56.73 & 58.12 & 61.40 & 58.60 & 67.61 & 59.05 & 62.97 \\
\quad\textit{w/ LLM-as-a-Judge RL} & 64.17 & 66.80 & 64.88 & 60.78 & 61.69 & 66.99 & 63.89 & 63.33 & 66.15 & 64.81 & 73.67 & 64.38 & 65.63 \\
\quad\textit{w/ ACE-RL} & \textbf{73.65} & \textbf{75.76} & \textbf{74.88} & \textbf{70.09} & \textbf{71.49} & \textbf{75.21} & \textbf{74.50} & \textbf{72.20} & \textbf{73.10} & \textbf{73.91} & \textbf{80.82} & \textbf{72.00} & \textbf{74.73} \\
\midrule

Qwen2.5-3B-instruct & 54.07 & 56.30 & 54.38 & 53.65 & 46.03 & 58.00 & 56.05 & 53.72 & 58.35 & 54.62 & 66.79 & 52.40 & 52.73 \\
\quad\textit{w/ LongWriter SFT} & 56.31 & 59.02 & 58.18 & 56.99 & 52.11 & 56.90 & 54.67 & 56.29 & 59.34 & 56.88 & 63.60 & 54.87 & 53.78 \\
\quad\textit{w/ LLM-as-a-Judge RL} & 68.70 & 69.64 & 68.83 & 67.79 & 67.95 & 71.23 & 66.73 & 69.07 & 72.33 & 69.24 & 77.40 & 68.02 & 69.32 \\
\quad\textit{w/ ACE-RL} & \textbf{78.97} & \textbf{79.49} & \textbf{79.40} & \textbf{78.50} & \textbf{76.82} & \textbf{81.17} & \textbf{78.42} & \textbf{78.94} & \textbf{81.16} & \textbf{79.48} & \textbf{87.08} & \textbf{78.22} & \textbf{80.92} \\
\midrule

Qwen2.5-7B-instruct & 57.04 & 58.57 & 55.85 & 55.28 & 52.96 & 59.48 & 60.11 & 57.35 & 62.33 & 57.68 & 70.20 & 56.58 & 55.63 \\
\quad\textit{w/ LongWriter SFT} & 62.11 & 63.81 & 63.93 & 61.49 & 60.59 & 62.00 & 60.86 & 62.12 & 64.79 & 62.80 & 69.13 & 61.59 & 59.86 \\
\quad\textit{w/ LLM-as-a-Judge RL} & 74.45 & 75.78 & 74.92 & 71.65 & 72.34 & 78.45 & 73.58 & 73.98 & 76.73 & 74.47 & 78.87 & 72.65 & 71.40 \\
\quad\textit{w/ ACE-RL} & \textbf{81.44} & \textbf{82.83} & \textbf{81.43} & \textbf{79.62} & \textbf{81.17} & \textbf{83.39} & \textbf{80.20} & \textbf{81.16} & \textbf{83.65} & \textbf{81.79} & \textbf{88.01} & \textbf{79.83} & \textbf{79.84} \\
\midrule

Qwen3-4B & 67.42 & 68.13 & 68.49 & 67.69 & 62.45 & 69.37 & 68.38 & 67.90 & 73.69 & 68.54 & 81.36 & 66.04 & 71.76 \\
\quad\textit{w/ LongWriter SFT} & 65.48 & 66.40 & 66.14 & 64.29 & 65.29 & 67.15 & 63.59 & 66.19 & 69.13 & 66.37 & 74.32 & 64.52 & 63.31 \\
\quad\textit{w/ LLM-as-a-Judge RL} & 77.71 & 76.02 & 76.99 & 77.28 & 78.25 & 79.14 & 78.56 & 78.75 & 84.17 & 78.01 & 85.07 & 77.78 & 77.23 \\
\quad\textit{w/ ACE-RL} & \textbf{83.24} & \textbf{83.21} & \textbf{83.36} & \textbf{82.54} & \textbf{84.03} & \textbf{84.04} & \textbf{82.27} & \textbf{83.76} & \textbf{87.16} & \textbf{84.07} & \textbf{89.82} & \textbf{82.71} & \textbf{85.61} \\
\midrule

Qwen3-8B & 70.75 & 70.68 & 70.74 & 70.33 & 68.51 & 72.74 & 71.52 & 71.72 & 76.77 & 71.69 & 83.12 & 70.06 & 75.36 \\
\quad\textit{w/ LongWriter SFT} & 66.44 & 67.15 & 66.74 & 65.86 & 66.71 & 67.19 & 65.00 & 66.79 & 69.47 & 67.34 & 74.70 & 65.42 & 62.77 \\
\quad\textit{w/ LLM-as-a-Judge RL} & 78.45 & 76.92 & 78.04 & 77.26 & 80.99 & 78.14 & 79.34 & 79.29 & 84.29 & 79.07 & 86.11 & 79.33 & 79.86 \\
\quad\textit{w/ ACE-RL} & \textbf{84.22} & \textbf{84.01} & \textbf{84.26} & \textbf{83.68} & \textbf{86.17} & \textbf{83.75} & \textbf{83.44} & \textbf{84.63} & \textbf{87.45} & \textbf{84.84} & \textbf{90.11} & \textbf{84.74} & \textbf{86.58} \\
\bottomrule
\end{tabular}}
\caption{Performance of different LLMs on WritingBench across six domains and three writing requirements. Scores are normalized from a 0-10 range to a 100-point scale. The corresponding domains and requirements are: (D1) Academic \& Engineering, (D2) Finance \& Business, (D3) Politics \& Law, (D4) Literature \& Art, (D5) Education, (D6) Advertising \& Marketing, (R1) Style, (R2) Format, and (R3) Length. "C" denotes the category-specific scores of the three requirements.}
\label{tab:main table}
\end{table*}

\subsection{Baselines}
We compare ACE-RL with the following methods:

\paragraph{Current Leading LLMs} We evaluate: (1) Proprietary LLMs: o3-2025-04-16 \citep{o3}, Gemini-2.5-pro-preview-06-05 \citep{comanici2025gemini}, Claude-3.7-sonnet \citep{claude-3-7}, GPT-4o \citep{GPT-4o}, o1-Preview \citep{o1}. (2) Open-source LLMs: Deepseek-R1 \citep{guo2025deepseek}, Qwen3 series LLMs \citep{yang2025qwen3}, Qwen2.5 series LLMs \citep{yang2024qwen2}, Llama3 series LLMs \cite{dubey2024llama}. (3) Writing-enhanced LLMs: LongWriter-llama3.1-8B, LongWriter-glm-9B \citep{bai2025longwriter}, LongWriter-Zero-32B \citep{wu2025longwriter}.

\paragraph{Long-form Generation SFT} We perform SFT on LongWriter dataset following \citet{bai2025longwriter}. 

\paragraph{LLM-as-a-Judge RL} Since our method does not require pairwise preference data for training, we adapt the LLM-as-a-Judge prompt (removing pair comparison) from \cite{lei2025writing} for RL training. To ensure a fair comparison, the same model is employed for LLM-as-a-Judge as is used in our ACE-RL constraint verification.

\subsection{Evaluation Benchmarks}
\paragraph{WritingBench} contains 1000 diverse long-form writing instructions over 6 major domains and 100 sub-domains \citep{wu2025writingbench}. Each is paired with several evaluation criteria of three dimensions: style, format, and length. The evaluation is conducted by a fine-tuned critic LLM, achieving over 80\% agreement with human judgments.

\paragraph{Arena-Write} comprises 100 real-world user writing queries with six strong baseline responses for comparison \cite{wu2025longwriter}: DeepSeek-R1 \citep{guo2025deepseek}, Qwen2.5-72B \citep{yang2024qwen2}, GPT-4o \citep{GPT-4o}, GLM-Z1-32B \citep{glm2024chatglm}, Claude-3.5-Sonnet \citep{claude-3-5}, and DeepSeek-R1-Distill-Qwen32B \citep{guo2025deepseek}, resulting in a total of 595 comparisons. We follow all its evaluation settings with Qwen3-235B-A22B as the judge model and calculate the winning rate of model responses against these six baselines.

\section{Results and Analysis}
We assess ACE-RL with various LLMs: Llama3.1-8B \citep{dubey2024llama}, Qwen2.5-3B/7B \citep{yang2024qwen2}, and Qwen3-4B/8B \citep{yang2025qwen3}. All models are evaluated with a non-thinking mode, unless explicitly stated.

\subsection{Main Results}
As illustrated in Table \ref{tab:main table}, we provide the detailed performance of diverse LLMs over WritingBench. On average, LLMs trained with our ACE-RL framework surpass the original ones, SFT, and LLM-as-a-Judge RL by 21.03\%, 18.63\%, and 7.61\%. These substantial improvements underscore the critical role of our instruction-adaptive reward mechanism in boosting the long-form writing capabilities of LLMs. We present cases of model responses in Appendix \ref{res}.

Notably, our method proves highly competitive even when compared with leading proprietary systems and specialized writing models. Our Qwen3-8B model (\textbf{84.22}) not only outperforms proprietary LLMs like Gemini-2.5-pro-preview (83.05) and GPT-4o (74.46), but also surpasses writing-enhanced models like LongWriter-Zero-32B (80.30), LongWriter-llama3.1-8B (58.01), and LongWriter-glm-9B (62.94). Moreover, we discover that model scale is not the primary determinant of long-form generation quality. Even our trained Qwen2.5-3B-instruct models (\textbf{78.97}) can achieve higher performance than Qwen2.5-72B-instruct (65.28) and Llama-3.3-70b-instruct (50.43), suggesting that a better training paradigm may be more critical to unlock better long-form generation performance.

To further validate our method across different evaluation paradigms, we report the pairwise "win-tie-loss" rate of our models against six baselines on the Arena-Write benchmark. As shown in Table \ref{tab:arena}, our method demonstrates a decisive advantage in preference evaluations. Our trained Qwen3-8B model achieves a remarkable win rate of 77.39\% when compared with six strong baselines, including larger-scale models or proprietary systems, underscoring the success of our ACE-RL framework in unlocking the long-form generation capability of LLMs.

\subsection{Superiority of RL over SFT}
A consistent trend across all model scales is the superior performance of RL-based methods over SFT-based methods. Notably, even the Qwen2.5-3B model trained with ACE-RL can achieve an average score of 78.97, significantly surpassing the 62.11 score of the much larger Qwen2.5-7B model with LongWriter SFT. This result underscores the critical role of the training paradigm in long-form generation. Moreover, for Qwen3 series LLMs, we observe a decline in performance after SFT, indicating that the effectiveness of SFT is inherently constrained by the quality of the training dataset. Any shortcomings or biases in the SFT data, such as a domain gap between the synthetic examples and the complexity of real-world scenarios, would directly constrain the model's performance ceiling. In contrast, RL-based training allows the model to move beyond simple imitation. By learning a policy through exploration and reward, LLMs develop a more robust and generalizable understanding of the task, enabling them to generate higher-quality outputs that are better aligned with complex, real-world objectives.

\begin{table}[!t]
    \centering
    \resizebox{1.0\columnwidth}{!}{\begin{tabular}{lccc}
        \toprule
        \textbf{Models} & \textbf{Win Rate} & \textbf{Tie Rate} & \textbf{Loss Rate} \\
        \midrule
        Llama-3.1-8B-instruct & 7.82 & 0.42 & 91.76 \\
        \quad\textit{w/ LongWriter SFT} & 38.82 & 0.50 & 60.68 \\
        \quad\textit{w/ LLM-as-a-Judge RL} & 29.08 & 0.50 & 70.42 \\
        \quad\textit{w/ ACE-RL} & \textbf{42.86} & 0.76 & 56.38 \\
        \midrule
        Qwen2.5-3B-instruct & 7.98 & 0.17 & 91.85 \\
        \quad\textit{w/ LongWriter SFT} & 30.34 & 0.25 & 69.41 \\
        \quad\textit{w/ LLM-as-a-Judge RL} & 33.53 & 0.50 & 65.97 \\
        \quad\textit{w/ ACE-RL} & \textbf{57.58} & 0.34 & 42.08 \\
        \midrule
        Qwen2.5-7B-instruct & 15.55 & 0.17 & 84.28 \\
        \quad\textit{w/ LongWriter SFT} & 41.93 & 0.42 & 57.65 \\
        \quad\textit{w/ LLM-as-a-Judge RL} & 43.70 & 0.67 & 55.63 \\
        \quad\textit{w/ ACE-RL} & \textbf{71.51} & 0.34 & 28.15 \\
        \midrule
        Qwen3-4B & 22.44 & 0.42 & 77.14 \\
        \quad\textit{w/ LongWriter SFT} & 42.27 & 0.34 & 57.39 \\
        \quad\textit{w/ LLM-as-a-Judge RL} & 61.34 & 0.34 & 38.32 \\
        \quad\textit{w/ ACE-RL} & \textbf{73.53} & 0.42 & 26.05 \\
        \midrule
        Qwen3-8B & 31.93 & 0.92 & 67.25 \\
        \quad\textit{w/ LongWriter SFT} & 44.79 & 0.25 & 54.96 \\
        \quad\textit{w/ LLM-as-a-Judge RL} & 66.89 & 0.34 & 32.77  \\
        \quad\textit{w/ ACE-RL} & \textbf{77.39} & 0.42 & 22.19 \\

        \bottomrule
    \end{tabular}}
    \caption{The win, tie, and loss rate of different methods against six strong baselines on Arena-Write.}
    \label{tab:arena}
\end{table}

\subsection{Effectiveness of Test-Time Scaling}

\begin{table}[!ht]
\centering
\footnotesize
\resizebox{1.0\columnwidth}{!}{
\begin{tabular}{l|c|cccccc}
\toprule
\multirow{2.5}{*}{Models} & \multirow{2.5}{*}{\textbf{Avg}} & \multicolumn{6}{c}{\textbf{Domains}} \\
 \cmidrule(lr){3-8}
& & D1 & D2 & D3 & D4 & D5 & D6 \\ 
\midrule
Qwen3-8B & 70.75 & 70.68 & 70.74 & 70.33 & 68.51 & 72.74 & 71.52 \\
\quad\textit{w/ ACE-RL} & 84.22 & 84.01 & 84.26 & 83.68 & 86.17 & 83.75 & 83.44 \\
Qwen3-8B-thinking & 75.56 & 75.47 & 74.42 & 75.51 & 73.03 & 77.80 & 77.16 \\
\quad\textit{w/ ACE-RL} & \textbf{85.28} & \textbf{85.05} & \textbf{84.63} & \textbf{84.20} & \textbf{86.58} & \textbf{85.98} & \textbf{85.23} \\
\bottomrule
\end{tabular}}
\caption{Performance comparison of thinking and non-thinking modes on WritingBench.}
\label{tab:tts}
\end{table}

Test-time scaling has been demonstrated to improve performance on reasoning tasks \citep{guo2025deepseek}. To investigate whether these benefits extend to long-form generation, we conduct comparative experiments on Qwen3-8B using two modes: "thinking" and "non-thinking". In the "thinking" mode, our training paradigm guides the model to first analyze the user's intent and then create a structured plan or chain of thought before generating the final response. In contrast, the "non-thinking" mode involves direct, single-pass generation of the response. The results, presented in Table \ref{tab:tts}, indicate that the explicit internal reasoning process leads to the generation of higher-quality content, thereby yielding significantly better performance than the "non-thinking" approach. We present cases of thinking processes and responses in Appendix \ref{res}.


\subsection{Effectiveness of Reward Design}
\begin{figure}[h!]
  \includegraphics[width=\linewidth]{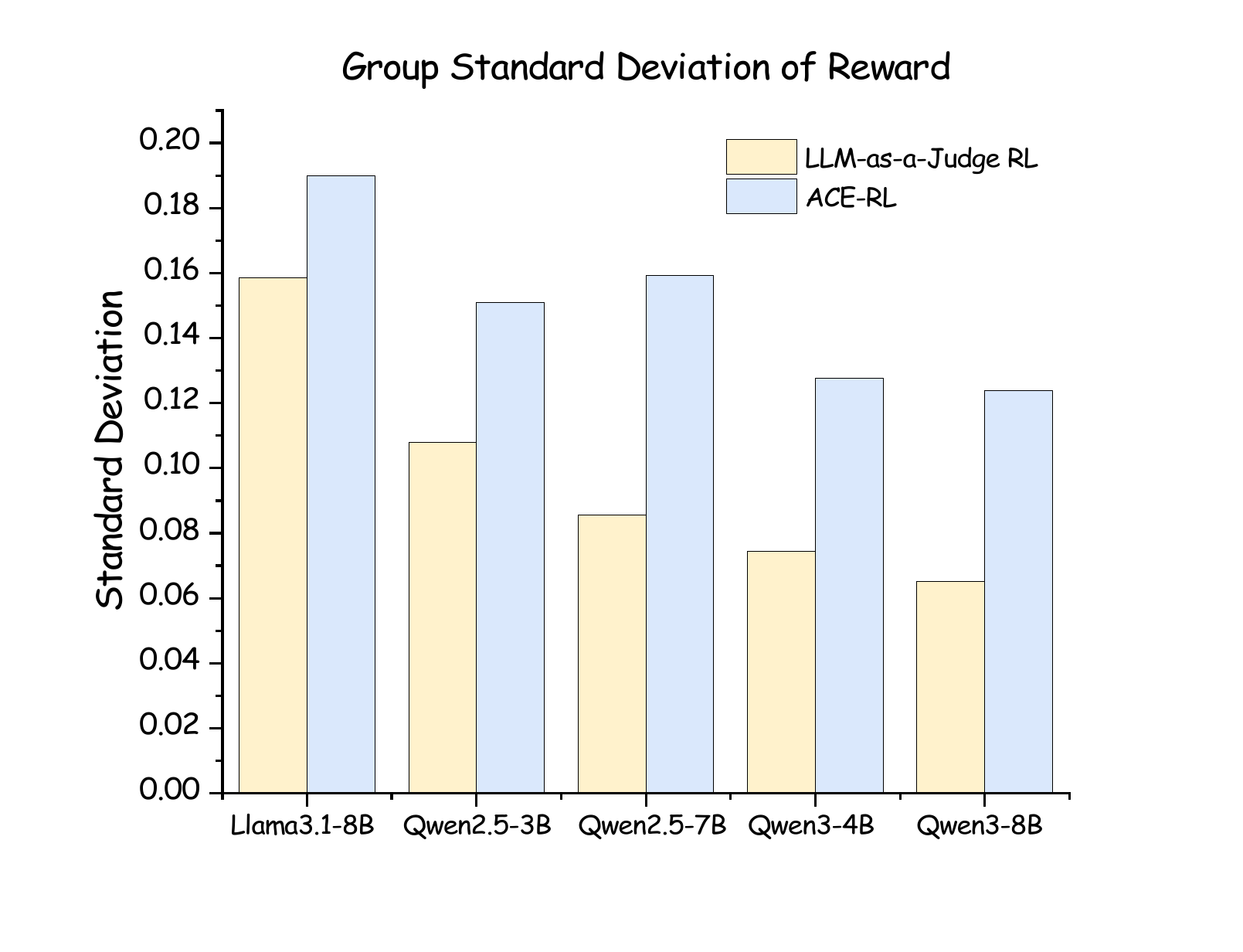}
  \caption{The comparison of the average group standard deviation of reward value.}
  \label{std}
\end{figure}
While LLM-as-a-Judge is a promising approach for guiding long-form generation, current methodologies present significant challenges in terms of cost, scalability, and reliability. For example, \citet{wu2025longwriter} train a Qwen2.5-72B reward model on manually-labeled preference data pairs, and \citet{lei2025writing} utilize Qwen-Plus \citep{yang2024qwen2} for pairwise comparison during RL training. Such methods introduce substantial cost and limit scalability due to dependency on additional labeled data. Furthermore, recent work shows that LLMs' judgment exhibits preference biases, causing the reward for both high-quality and low-quality responses to cluster within a narrow range \citep{li2025evaluating}. This lack of discriminatory power is evident in Figure \ref{std}, where the reward distribution within groups from an LLM-as-a-Judge shows substantially lower variance than that of our ACE-RL method. This suggests that LLMs' judgment based on general dimensions like coherence, relevance, and helpfulness can not distinguish responses of different quality. In contrast, our method converts the coarse-grained, subjective evaluation of long-form responses into fine-grained constraint verification tasks, enhancing the efficiency and reliability of reward generation with small-scale models.


%

\subsection{Training Data Quality Analysis}
\label{data_quality}
\begin{table}[!ht]
\centering
\footnotesize
\resizebox{1.0\columnwidth}{!}{
\begin{tabular}{l|c|cccccc}
\toprule
\multirow{2.5}{*}{Method} & \multirow{2.5}{*}{\textbf{Avg}} & \multicolumn{6}{c}{\textbf{Domains}} \\
 \cmidrule(lr){3-8}
& & D1 & D2 & D3 & D4 & D5 & D6 \\ 
\midrule
ACE-RL & \textbf{81.44} & \textbf{82.83} & \textbf{81.43} & \textbf{79.62} & \textbf{81.17} & \textbf{83.39} & \textbf{80.20} \\
\quad\textit{w/o Filter} & 78.25 & 79.72 & 77.89 & 77.50 & 77.45 & 80.50 & 76.47 \\
\bottomrule
\end{tabular}}
\caption{Performance comparison of Qwen2.5-7B on WritingBench with and without the data quality filtering strategy.}
\label{tab:filter}
\end{table}

To ensure the quality of training data, we conduct an analysis of instruction–response pairs based on the constraint-based verification scores generated by Qwen3-235B-A22B. Each instruction is paired with eight diverse responses sampled from Qwen2.5-7B, and each response is evaluated against the corresponding constraint checklist to produce a verification score in $\{0, 0.5, 1\}$. The average of these eight scores serves as a proxy for the intrinsic difficulty of the instruction: a high average suggests that nearly all sampled responses satisfy the constraints, indicating either overly permissive constraints or a trivial instruction that fails to challenge the model. Including such examples in RL training introduces minimal signal, as the reward becomes identical, failing to differentiate between high and low quality behaviors. As shown in Table \ref{tab:filter}, applying this filtering strategy yields a consistent performance boost across all six domains of the WritingBench benchmark, outperforming the variant without filtering by 3.19\%. This demonstrates that more challenging data can enhance the efficiency and effectiveness of reinforcement learning by ensuring richer reward signals during training.

\subsection{Self-Rewarding Enables Effective Bootstrapping}
In a fully self-bootstrapping setting, where the Qwen3-8B model generates and learns from its own reward signals, it achieves a score of 84.22. This demonstrates that ACE-RL enables a model to effectively act as its own critic without the need for larger or proprietary reward models. This capability stems from the fundamental asymmetry between generation and verification: checking for concrete constraints is significantly simpler than the open-ended long-form generation. Consequently, ACE-RL mitigates the need for expensive external supervision, paving the way for scalable, self-evolving training pipelines where models can refine their performance through self-rewarding.

\subsection{Human Evaluation}

\begin{figure}[h!]
  \includegraphics[width=\linewidth]{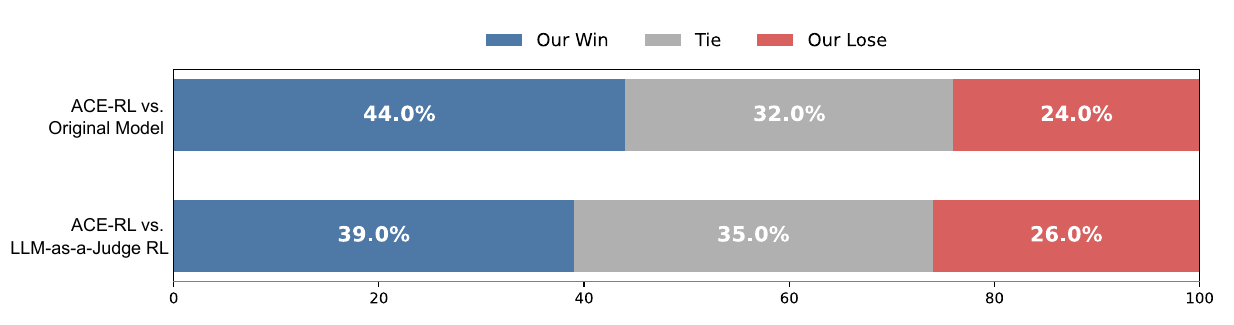}
  \caption{Human preference evaluation between our ACE-RL method and different baselines.}
  \label{human}
\end{figure}

To mitigate biases inherent in automated evaluation, we conduct human evaluation with real-world instructions from the Arena-Write benchmark. We compared Qwen-3-8B trained with ACE-RL against its base model and the LLM-as-a-Judge RL method. As shown in Figure \ref{human}, our method consistently achieves higher win rates, demonstrating its superiority in aligning with human preferences.

\section{Conclusion}
This paper introduces ACE-RL, a novel and efficient reinforcement learning framework to enhance the long-form generation capability of LLMs. ACE-RL leverages an effective reward mechanism that measures the quality of long-form responses via instruction-adaptive constraints verification. By converting coarse-grained, subjective evaluation to fine-grained verification tasks, LLMs trained with ACE-RL demonstrate their superiority over other SFT and RL methods, and even surpass leading proprietary LLMs. ACE-RL offers new pathways for a verifiable reward-based RL paradigm for long-form generation scenarios.


\bibliography{tacl2021}

@misc{comanici2025gemini25pushingfrontier,
      title={Gemini 2.5: Pushing the Frontier with Advanced Reasoning, Multimodality, Long Context, and Next Generation Agentic Capabilities}, 
      author={Google},
      year={2025},
      eprint={2507.06261},
      archivePrefix={arXiv},
      primaryClass={cs.CL},
      url={https://arxiv.org/abs/2507.06261}, 
}

@Misc{o3,
title = {OpenAI o3},
author={OpenAI},
url={https://platform.openai.com/docs/models/o3},
year = {2025}
}

@Misc{o1,
title = {OpenAI o1},
author={OpenAI},
url={https://platform.openai.com/docs/models/o1},
year = {2024}
}

@misc{claude-4,
  author = {Anthropic},
  title = {Anthropic: Introducing Claude 4},
  year = {2025},
  url = {https://www.anthropic.com/news/claude-4},
}

@article{yang2025qwen3,
  title={Qwen3 technical report},
  author={Yang, An and Li, Anfeng and Yang, Baosong and Zhang, Beichen and Hui, Binyuan and Zheng, Bo and Yu, Bowen and Gao, Chang and Huang, Chengen and Lv, Chenxu and others},
  journal={arXiv preprint arXiv:2505.09388},
  year={2025}
}

@inproceedings{
wu2025longgenbench,
title={LongGenBench: Benchmarking Long-Form Generation in Long Context {LLM}s},
author={Yuhao Wu and Ming Shan Hee and Zhiqiang Hu and Roy Ka-Wei Lee},
booktitle={The Thirteenth International Conference on Learning Representations},
year={2025},
url={https://openreview.net/forum?id=3A71qNKWAS}
}

@article{wu2025shifting,
  title={Shifting long-context llms research from input to output},
  author={Wu, Yuhao and Bai, Yushi and Hu, Zhiqing and Tu, Shangqing and Hee, Ming Shan and Li, Juanzi and Lee, Roy Ka-Wei},
  journal={arXiv preprint arXiv:2503.04723},
  year={2025}
}

@inproceedings{
bai2025longwriter,
title={LongWriter: Unleashing 10,000+ Word Generation from Long Context {LLM}s},
author={Yushi Bai and Jiajie Zhang and Xin Lv and Linzhi Zheng and Siqi Zhu and Lei Hou and Yuxiao Dong and Jie Tang and Juanzi Li},
booktitle={The Thirteenth International Conference on Learning Representations},
year={2025},
url={https://openreview.net/forum?id=kQ5s9Yh0WI}
}

@article{wu2025writingbench,
  title={Writingbench: A comprehensive benchmark for generative writing},
  author={Wu, Yuning and Mei, Jiahao and Yan, Ming and Li, Chenliang and Lai, Shaopeng and Ren, Yuran and Wang, Zijia and Zhang, Ji and Wu, Mengyue and Jin, Qin and others},
  journal={arXiv preprint arXiv:2503.05244},
  year={2025}
}

@article{quan2024language,
  title={Language models can self-lengthen to generate long texts},
  author={Quan, Shanghaoran and Tang, Tianyi and Yu, Bowen and Yang, An and Liu, Dayiheng and Gao, Bofei and Tu, Jianhong and Zhang, Yichang and Zhou, Jingren and Lin, Junyang},
  journal={arXiv preprint arXiv:2410.23933},
  year={2024}
}

@inproceedings{pham2024suri,
  title={Suri: Multi-constraint Instruction Following in Long-form Text Generation},
  author={Pham, Chau and Sun, Simeng and Iyyer, Mohit},
  booktitle={Findings of the Association for Computational Linguistics: EMNLP 2024},
  pages={1722--1753},
  year={2024}
}

@inproceedings{deng2022model,
  title={Model Criticism for Long-Form Text Generation},
  author={Deng, Yuntian and Kuleshov, Volodymyr and Rush, Alexander M},
  booktitle={Proceedings of the 2022 Conference on Empirical Methods in Natural Language Processing},
  pages={11887--11912},
  year={2022}
}

@article{lei2025writing,
  title={Writing-RL: Advancing Long-form Writing via Adaptive Curriculum Reinforcement Learning},
  author={Lei, Xuanyu and Li, Chenliang and Wu, Yuning and Liu, Kaiming and Shen, Weizhou and Li, Peng and Yan, Ming and Zhang, Ji and Huang, Fei and Liu, Yang},
  journal={arXiv preprint arXiv:2506.05760},
  year={2025}
}

@article{wu2025longwriter,
  title={LongWriter-Zero: Mastering Ultra-Long Text Generation via Reinforcement Learning},
  author={Wu, Yuhao and Bai, Yushi and Hu, Zhiqiang and Lee, Roy Ka-Wei and Li, Juanzi},
  journal={arXiv preprint arXiv:2506.18841},
  year={2025}
}

@article{shao2024deepseekmath,
  title={Deepseekmath: Pushing the limits of mathematical reasoning in open language models},
  author={Shao, Zhihong and Wang, Peiyi and Zhu, Qihao and Xu, Runxin and Song, Junxiao and Bi, Xiao and Zhang, Haowei and Zhang, Mingchuan and Li, YK and Wu, Y and others},
  journal={arXiv preprint arXiv:2402.03300},
  year={2024}
}

@inproceedings{
zhao2024wildchat,
title={WildChat: 1M Chat{GPT} Interaction Logs in the Wild},
author={Wenting Zhao and Xiang Ren and Jack Hessel and Claire Cardie and Yejin Choi and Yuntian Deng},
booktitle={The Twelfth International Conference on Learning Representations},
year={2024},
url={https://openreview.net/forum?id=Bl8u7ZRlbM}
}

@article{liu2025comprehensive,
  title={A comprehensive survey on long context language modeling},
  author={Liu, Jiaheng and Zhu, Dawei and Bai, Zhiqi and He, Yancheng and Liao, Huanxuan and Que, Haoran and Wang, Zekun and Zhang, Chenchen and Zhang, Ge and Zhang, Jiebin and others},
  journal={arXiv preprint arXiv:2503.17407},
  year={2025}
}

@inproceedings{
jin2025longcontext,
title={Long-Context {LLM}s Meet {RAG}: Overcoming Challenges for Long Inputs in {RAG}},
author={Bowen Jin and Jinsung Yoon and Jiawei Han and Sercan O Arik},
booktitle={The Thirteenth International Conference on Learning Representations},
year={2025},
url={https://openreview.net/forum?id=oU3tpaR8fm}
}

@inproceedings{
gu2024mamba,
title={Mamba: Linear-Time Sequence Modeling with Selective State Spaces},
author={Albert Gu and Tri Dao},
booktitle={First Conference on Language Modeling},
year={2024},
url={https://openreview.net/forum?id=tEYskw1VY2}
}

@inproceedings{
dao2024transformers,
title={Transformers are {SSM}s: Generalized Models and Efficient Algorithms Through Structured State Space Duality},
author={Tri Dao and Albert Gu},
booktitle={Forty-first International Conference on Machine Learning},
year={2024},
url={https://openreview.net/forum?id=ztn8FCR1td}
}

@inproceedings{peng2023rwkv,
  title={RWKV: Reinventing RNNs for the Transformer Era},
  author={Peng, Bo and Alcaide, Eric and Anthony, Quentin and Albalak, Alon and Arcadinho, Samuel and Biderman, Stella and Cao, Huanqi and Cheng, Xin and Chung, Michael and Derczynski, Leon and others},
  booktitle={Findings of the Association for Computational Linguistics: EMNLP 2023},
  pages={14048--14077},
  year={2023}
}

@article{lu2025moba,
  title={Moba: Mixture of block attention for long-context llms},
  author={Lu, Enzhe and Jiang, Zhejun and Liu, Jingyuan and Du, Yulun and Jiang, Tao and Hong, Chao and Liu, Shaowei and He, Weiran and Yuan, Enming and Wang, Yuzhi and others},
  journal={arXiv preprint arXiv:2502.13189},
  year={2025}
}

@article{shah2024flashattention,
  title={Flashattention-3: Fast and accurate attention with asynchrony and low-precision},
  author={Shah, Jay and Bikshandi, Ganesh and Zhang, Ying and Thakkar, Vijay and Ramani, Pradeep and Dao, Tri},
  journal={Advances in Neural Information Processing Systems},
  volume={37},
  pages={68658--68685},
  year={2024}
}

@inproceedings{xiong2024effective,
  title={Effective Long-Context Scaling of Foundation Models},
  author={Xiong, Wenhan and Liu, Jingyu and Molybog, Igor and Zhang, Hejia and Bhargava, Prajjwal and Hou, Rui and Martin, Louis and Rungta, Rashi and Sankararaman, Karthik Abinav and Oguz, Barlas and others},
  booktitle={Proceedings of the 2024 Conference of the North American Chapter of the Association for Computational Linguistics: Human Language Technologies (Volume 1: Long Papers)},
  pages={4643--4663},
  year={2024}
}

@inproceedings{
peng2024yarn,
title={Ya{RN}: Efficient Context Window Extension of Large Language Models},
author={Bowen Peng and Jeffrey Quesnelle and Honglu Fan and Enrico Shippole},
booktitle={The Twelfth International Conference on Learning Representations},
year={2024},
url={https://openreview.net/forum?id=wHBfxhZu1u}
}

@inproceedings{
ding2024longrope,
title={LongRo{PE}: Extending {LLM} Context Window Beyond 2 Million Tokens},
author={Yiran Ding and Li Lyna Zhang and Chengruidong Zhang and Yuanyuan Xu and Ning Shang and Jiahang Xu and Fan Yang and Mao Yang},
booktitle={Forty-first International Conference on Machine Learning},
year={2024},
url={https://openreview.net/forum?id=ONOtpXLqqw}
}

@inproceedings{
an2024trainingfree,
title={Training-Free Long-Context Scaling of Large Language Models},
author={Chenxin An and Fei Huang and Jun Zhang and Shansan Gong and Xipeng Qiu and Chang Zhou and Lingpeng Kong},
booktitle={Forty-first International Conference on Machine Learning},
year={2024},
url={https://openreview.net/forum?id=If4xW9vF7U}
}

@inproceedings{fu2024data,
    title={Data Engineering for Scaling Language Models to 128K Context},
    author={Yao Fu and Rameswar Panda and Xinyao Niu and Xiang Yue and Hannaneh Hajishirzi and Yoon Kim and Hao Peng},
    booktitle={Forty-first International Conference on Machine Learning},
    year={2024},
    url={https://openreview.net/forum?id=TaAqeo7lUh}
}

@inproceedings{chen2024long,
  title={Long Context is Not Long at All: A Prospector of Long-Dependency Data for Large Language Models},
  author={Chen, Longze and Liu, Ziqiang and He, Wanwei and Zheng, Yinhe and Sun, Hao and Li, Yunshui and Luo, Run and Yang, Min},
  booktitle={Proceedings of the 62nd Annual Meeting of the Association for Computational Linguistics (Volume 1: Long Papers)},
  pages={8222--8234},
  year={2024}
}

@inproceedings{chen2025ladm,
    title = "{LADM}: Long-context Training Data Selection with Attention-based Dependency Measurement for {LLM}s",
    author = "Chen, Jianghao  and
      Wu, Junhong  and
      Xu, Yangyifan  and
      Zhang, Jiajun",
    editor = "Che, Wanxiang  and
      Nabende, Joyce  and
      Shutova, Ekaterina  and
      Pilehvar, Mohammad Taher",
    booktitle = "Proceedings of the 63rd Annual Meeting of the Association for Computational Linguistics (Volume 1: Long Papers)",
    month = jul,
    year = "2025",
    address = "Vienna, Austria",
    publisher = "Association for Computational Linguistics",
    url = "https://aclanthology.org/2025.acl-long.154/",
    doi = "10.18653/v1/2025.acl-long.154",
    pages = "3076--3090",
    ISBN = "979-8-89176-251-0"
}

@inproceedings{bai2024longalign,
  title={LongAlign: A Recipe for Long Context Alignment of Large Language Models},
  author={Bai, Yushi and Lv, Xin and Zhang, Jiajie and He, Yuze and Qi, Ji and Hou, Lei and Tang, Jie and Dong, Yuxiao and Li, Juanzi},
  booktitle={Findings of the Association for Computational Linguistics: EMNLP 2024},
  pages={1376--1395},
  year={2024}
}

@article{an2024make,
  title={Make your llm fully utilize the context},
  author={An, Shengnan and Ma, Zexiong and Lin, Zeqi and Zheng, Nanning and Lou, Jian-Guang and Chen, Weizhu},
  journal={Advances in Neural Information Processing Systems},
  volume={37},
  pages={62160--62188},
  year={2024}
}

@inproceedings{
chen2024longlora,
title={LongLo{RA}: Efficient Fine-tuning of Long-Context Large Language Models},
author={Yukang Chen and Shengju Qian and Haotian Tang and Xin Lai and Zhijian Liu and Song Han and Jiaya Jia},
booktitle={The Twelfth International Conference on Learning Representations},
year={2024},
url={https://openreview.net/forum?id=6PmJoRfdaK}
}

@article{comanici2025gemini,
  title={Gemini 2.5: Pushing the frontier with advanced reasoning, multimodality, long context, and next generation agentic capabilities},
  author={Comanici, Gheorghe and Bieber, Eric and Schaekermann, Mike and Pasupat, Ice and Sachdeva, Noveen and Dhillon, Inderjit and Blistein, Marcel and Ram, Ori and Zhang, Dan and Rosen, Evan and others},
  journal={arXiv preprint arXiv:2507.06261},
  year={2025}
}

@misc{claude-3-7,
  author = {Anthropic},
  title = {Anthropic: Introducing Claude 3.7 Sonnet},
  year = {2024},
  url = {https://www.anthropic.com/news/claude-3-7-sonnet},
}

@misc{claude-3-5,
  author = {Anthropic},
  title = {Anthropic: Introducing Claude 3.5 Sonnet},
  year = {2024},
  url = {https://www.anthropic.com/news/claude-3-5-sonnet},
}

@misc{GPT-4o,
  author = {OpenAI},
  title = {OpenAI: Hello GPT-4o},
  year = {2024},
  url = {https://openai.com/index/hello-gpt-4o/},
}

@article{guo2025deepseek,
  title={Deepseek-r1: Incentivizing reasoning capability in llms via reinforcement learning},
  author={Guo, Daya and Yang, Dejian and Zhang, Haowei and Song, Junxiao and Zhang, Ruoyu and Xu, Runxin and Zhu, Qihao and Ma, Shirong and Wang, Peiyi and Bi, Xiao and others},
  journal={arXiv preprint arXiv:2501.12948},
  year={2025}
}

@article{liu2024deepseek,
  title={Deepseek-v3 technical report},
  author={Liu, Aixin and Feng, Bei and Xue, Bing and Wang, Bingxuan and Wu, Bochao and Lu, Chengda and Zhao, Chenggang and Deng, Chengqi and Zhang, Chenyu and Ruan, Chong and others},
  journal={arXiv preprint arXiv:2412.19437},
  year={2024}
}

@article{yang2024qwen2,
  title={Qwen2. 5 technical report},
  author={Yang, An and Yang, Baosong and Zhang, Beichen and Hui, Binyuan and Zheng, Bo and Yu, Bowen and Li, Chengyuan and Liu, Dayiheng and Huang, Fei and Wei, Haoran and others},
  journal={arXiv preprint arXiv:2412.15115},
  year={2024}
}

@article{dubey2024llama,
  title={The llama 3 herd of models},
  author={Dubey, Abhimanyu and Jauhri, Abhinav and Pandey, Abhinav and Kadian, Abhishek and Al-Dahle, Ahmad and Letman, Aiesha and Mathur, Akhil and Schelten, Alan and Yang, Amy and Fan, Angela and others},
  journal={arXiv preprint arXiv:2407.21783},
  year={2024}
}

@inproceedings{sheng2025hybridflow,
  title={Hybridflow: A flexible and efficient rlhf framework},
  author={Sheng, Guangming and Zhang, Chi and Ye, Zilingfeng and Wu, Xibin and Zhang, Wang and Zhang, Ru and Peng, Yanghua and Lin, Haibin and Wu, Chuan},
  booktitle={Proceedings of the Twentieth European Conference on Computer Systems},
  pages={1279--1297},
  year={2025}
}

@article{yu2025dapo,
  title={Dapo: An open-source llm reinforcement learning system at scale},
  author={Yu, Qiying and Zhang, Zheng and Zhu, Ruofei and Yuan, Yufeng and Zuo, Xiaochen and Yue, Yu and Dai, Weinan and Fan, Tiantian and Liu, Gaohong and Liu, Lingjun and others},
  journal={arXiv preprint arXiv:2503.14476},
  year={2025}
}

@article{glm2024chatglm,
  title={Chatglm: A family of large language models from glm-130b to glm-4 all tools},
  author={GLM, Team and Zeng, Aohan and Xu, Bin and Wang, Bowen and Zhang, Chenhui and Yin, Da and Zhang, Dan and Rojas, Diego and Feng, Guanyu and Zhao, Hanlin and others},
  journal={arXiv preprint arXiv:2406.12793},
  year={2024}
}

@article{li2025evaluating,
  title={Evaluating Scoring Bias in LLM-as-a-Judge},
  author={Li, Qingquan and Dou, Shaoyu and Shao, Kailai and Chen, Chao and Hu, Haixiang},
  journal={arXiv preprint arXiv:2506.22316},
  year={2025}
}

@article{schulman2017proximal,
  title={Proximal policy optimization algorithms},
  author={Schulman, John and Wolski, Filip and Dhariwal, Prafulla and Radford, Alec and Klimov, Oleg},
  journal={arXiv preprint arXiv:1707.06347},
  year={2017}
}

@inproceedings{sun2025rethinking,
  title={Rethinking reward modeling in preference-based large language model alignment},
  author={Sun, Hao and Shen, Yunyi and Ton, Jean-Francois},
  booktitle={The Thirteenth International Conference on Learning Representations},
  year={2025}
}

@inproceedings{yang-etal-2025-implicit,
    title = "Implicit Cross-Lingual Rewarding for Efficient Multilingual Preference Alignment",
    author = "Yang, Wen  and
      Wu, Junhong  and
      Wang, Chen  and
      Zong, Chengqing  and
      Zhang, Jiajun",
    editor = "Che, Wanxiang  and
      Nabende, Joyce  and
      Shutova, Ekaterina  and
      Pilehvar, Mohammad Taher",
    booktitle = "Findings of the Association for Computational Linguistics: ACL 2025",
    month = jul,
    year = "2025",
    address = "Vienna, Austria",
    publisher = "Association for Computational Linguistics",
    url = "https://aclanthology.org/2025.findings-acl.1088/",
    doi = "10.18653/v1/2025.findings-acl.1088",
    pages = "21125--21147",
    ISBN = "979-8-89176-256-5"
}

@article{christiano2017deep,
  title={Deep reinforcement learning from human preferences},
  author={Christiano, Paul F and Leike, Jan and Brown, Tom and Martic, Miljan and Legg, Shane and Amodei, Dario},
  journal={Advances in neural information processing systems},
  volume={30},
  year={2017}
}
\bibliographystyle{acl_natbib}



\clearpage

\appendix
\newcommand{\miniscule}{\fontsize{7}{8}\selectfont}
\section{Prompt Templates}
\label{prompt}
\subsection{Prompt for Data Filtering}
We employ the following prompt to filter out instructions that specifically call for long-form generation.
\begin{tcolorbox}[breakable, title=Data Filtering, left=5pt, right=5pt, fonttitle=\small]
\miniscule
\textbf{\# Objective} \\
You are an AI assistant tasked with analyzing user queries to predict the length of the generated response. Your goal is to determine if a query is likely to require a long-form answer. \\
\\
\textbf{\# Response Categories} \\
- Yes: The response is highly likely to exceed 2000 words. \\
- Maybe: The response is likely to be on the borderline, or it is difficult to determine with high confidence. \\
- No: The response is unlikely to exceed 2000 words. \\
\\
\textbf{\# Evaluation Criteria} \\
Analyze the Query below based on these five factors: \\
\textbf{1. Completeness of Instruction:} Does the query provide all necessary information for a meaningful response? \\
Yes Example: "Write a practical report on the causes of environmental pollution." \\
No Example: "Read and analyze this paper." (Insufficient context provided) \\
\\
\textbf{2. Depth \& Complexity:} Does the query require in-depth explanations, nuanced analysis, or discussion of complex, multi-faceted topics? \\
Yes Example: "Provide a detailed analysis of the socio-economic impacts of the Industrial Revolution, including long-term consequences." \\
No Example: "Who was the first president of the United States?" \\
\\
\textbf{3. Scope \& Breadth:} Does the query cover a broad field, multiple distinct sub-topics, or require a comparison of various perspectives? \\
Yes Example: "Write a comprehensive history of ancient Greece, covering its philosophy, art, politics, and warfare." \\
No Example: "What are the key differences between a Roth IRA and a traditional IRA?" \\
\\
\textbf{4. Structural Needs:} Would a comprehensive answer necessitate a formal structure with multiple distinct sections (e.g., introduction, background, analysis, conclusion, bibliography)? \\
Yes Example: "Draft a research paper on the efficacy of renewable energy sources in combating climate change." \\
No Example: "Explain how photosynthesis works in simple terms." \\
\\
\textbf{5. Research \& Synthesis:} Does the query imply the need to gather, synthesize, and cite information from multiple external sources? \\
Yes Example: "Compile a literature review on the current state of CRISPR-Cas9 gene-editing technology, including recent breakthroughs and ethical considerations."
No Example: "Summarize the plot of the movie 'Inception'." \\
\\
\textbf{\# Output Requirements:} \\
Analysis: Provide a concise, point-by-point justification for your decision by evaluating the Query against each of the four factors listed above. \\
Decision: Conclude with a definitive answer from the Response Categories, enclosed in <Answer> and </Answer> tags: \\
<Answer>Yes</Answer> \\
<Answer>Maybe</Answer> \\
<Answer>No</Answer> \\

Query: \\
\{QUERY\}
\end{tcolorbox}

\subsection{Prompt for Constraint Generation}
We employ the following prompt to construct a constraint checklist for a specific long-form generation instruction.

\begin{tcolorbox}[breakable, title=Constraint Generation, left=5pt, right=5pt, fonttitle=\small]
\miniscule
You are tasked with creating an evaluation checklist. This checklist will be used to assess an AI assistant's response to a specific INSTRUCTION. Your goal is to generate a series of questions that determine if the AI's response has met the criteria outlined in the INSTRUCTION, as well as general quality standards. \\
\\
\textbf{\# Task Details} \\
Analyze the INSTRUCTION: \\
Carefully review the provided INSTRUCTION that the AI assistant was meant to follow. If the INSTRUCTION is broad or lacks specific details, your analysis should identify common implicit expectations and potential areas where a high-quality response would add value beyond the literal instruction. When analyzing the INSTRUCTION, consider the following dimensions: \\
\textbf{1. Content Completeness:} Identify all explicit requirements, topics, sub-questions, and data points mentioned in the INSTRUCTION. Infer any implicit expectations that a comprehensive response should satisfy. \\
\textbf{2. Structural Logic:} Determine if the INSTRUCTION specifies any organizational flow (e.g., "compare and contrast," "step-by-step," "introduction, body, conclusion"). If no structure is specified, consider what a logical and coherent arrangement of ideas would be for the given task. \\
\textbf{3. Stylistic Formatting:} Note any stylistic or formatting preferences mentioned in the INSTRUCTION (e.g., "use bullet points," "formal tone," "concise language"). If none are specified, infer appropriate stylistic choices based on the context and purpose of the INSTRUCTION. \\
\textbf{4. Additional Quality Criteria:} Reflect on other general quality aspects that would enhance the AI's response, such as clarity, relevance, depth of analysis, creativity, and engagement. \\
\\
\textbf{Develop Checklist Questions:} \\
Based on your analysis, create a list of questions. These questions will form the evaluation checklist. The checklist should assess whether the AI's response meets criteria that are explicitly stated in the INSTRUCTION. Crucially, for broad INSTRUCTIONS, the checklist should also assess criteria that are implicitly relevant, generally sensible, or represent a valuable and comprehensive approach to the problem domain of the INSTRUCTION, even if not explicitly requested by the user. Strive for conciseness; only include questions that are clearly relevant and necessary for evaluation. \\
\\
\textbf{\# Guidelines for Checklist Questions} \\
Each question in your checklist must: \\
\textbf{1. Be answerable with a three-level rating.} The answer should indicate the degree to which the AI's response successfully met the specific criterion being assessed. Recommended ratings are: \\
- Fully Met: AI response completely and excellently satisfies the criterion. \\
- Partially Met: AI response partially satisfies the criterion, with noticeable strengths but also significant areas for improvement. \\
- Not Met: AI response completely fails to satisfy the criterion. \\
\textbf{2. Be comprehensive yet concise.} Ensure all critical aspects of the INSTRUCTION are covered, but avoid redundancy or overly granular questions. Aim for a balance that captures essential evaluation points efficiently. Typically, a checklist should contain between two and eight questions. \\
\textbf{3. Be precise and unambiguous.} Questions should use clear language. Where appropriate, directly reference or paraphrase phrasing from the INSTRUCTION to ensure clarity on what is being evaluated. For broad INSTRUCTIONS, articulate clear and valuable criteria that an excellent AI response would naturally include, even if not explicitly stated by the user. Avoid vague terms. \\
\textbf{4. Focus on a single aspect.} Each question should ideally evaluate one specific element of the response. \\
\\
\textbf{\# Response Format} \\
Your output must strictly follow this format: \\
Analysis: Provide a brief analysis of the INSTRUCTION here. This analysis should identify the key requirements and implicit expectations of the INSTRUCTION that will inform your checklist questions. \\
<Checklist> \\
Question 1: Phrased for a three-level rating. \\
Question 2: Phrased for a three-level rating. \\
And so on... \\
</Checklist> \\
\\
INSTRUCTION: \\
\{INSTRUCTION\}
\end{tcolorbox}

\subsection{Prompt for Length Requirement}
We employ the following prompt to augment each instruction with the target word count range.
\begin{tcolorbox}[breakable, title=Length Requirement, left=5pt, right=5pt, fonttitle=\small]
\miniscule
You are an expert AI performance analyst. Your task is to predict the optimal and acceptable length range for an AI's response to a given user INSTRUCTION, considering the detailed requirements outlined in its corresponding EVALUATION CHECKLIST. Your prediction should guide an AI to generate a response that is comprehensive without being verbose, and concise without omitting essential details. \\
\\
\textbf{\# Task Details} \\
Analyze the INSTRUCTION and the EVALUATION CHECKLIST: \\
1. First, examine the INSTRUCTION to determine if it already contains an explicit number of length (e.g., "in 500 words,"). This is your first point of analysis. \\
2. Carefully review the user INSTRUCTION to grasp its core request, scope, and any initial implied complexity. \\
3. Thoroughly examine each question in the EVALUATION CHECKLIST. Understand that each "Fully Met" criterion in the checklist represents a specific piece of information, level of detail, or structural element that the AI's response must contain to be considered high-quality. These criteria directly influence the necessary content volume. \\
4. Identify the cumulative informational burden: Consider how many distinct points, examples, explanations, or structured elements the AI needs to generate to satisfy all "Fully Met" criteria across the checklist. This assessment is crucial for predicting length. \\
\\
\textbf{\# Predict Optimal Length} \\
1. Determine the Acceptable Length Range: This broader range defines the minimum and maximum count within which the AI's response would still be considered a high-quality, effective, and complete answer, even if not absolutely perfect. Responses outside this range are likely to be either incomplete or unnecessarily verbose. If the instruction specifies a length, your prediction should reflect that. \\ 
2. For English INSTRUCTIONS, predict length in words. For Chinese INSTRUCTIONS, predict length in characters. \\
3. The minimum possible word/character count is 0. The maximum possible word/character count is 10000. All predicted upper and lower bounds must be multiples of 100. The difference between the upper and lower bounds for any given range must not exceed 2000 words/characters. \\
\\
\textbf{\# Response Format} \\
Your output must strictly follow this format: \\
Analysis: Briefly explain how the complexity of the INSTRUCTION and the specific requirements outlined in the EVALUATION CHECKLIST inform your length predictions. Highlight which checklist items are the primary drivers of response length. \\
<LENGTH> \\
{[}length lower bound, length upper bound{]} \\
</LENGTH> \\
\\
INSTRUCTION: \\
\{INSTRUCTION\} \\
\\
EVALUATION CHECKLIST: \\
\{CHECKLIST\}
\end{tcolorbox}

\subsection{Prompt for Constraint Verification}
We employ the following prompt for a verifier LLM to assess a response over the original user instruction and a given constraint checklist.

\begin{tcolorbox}[breakable, title=Constraint Verification, left=5pt, right=5pt, fonttitle=\small]
\miniscule
You are a meticulous AI Quality Analyst. Your role is to evaluate an AI-generated RESPONSE based on its adherence to a given INSTRUCTION. The evaluation must be performed against a CHECKLIST of criteria. \\
\\
\textbf{\# Provided Information} \\
INSTRUCTION: The original request given to the AI. \\
\{INSTRUCTION\} \\
\\
RESPONSE: The AI's generated output that you must evaluate. \\
\{RESPONSE\} \\
\\
CHECKLIST: The list of criteria the RESPONSE must satisfy. \\
\{CHECKLIST\} \\
\\
\textbf{\# Task \& Instructions} \\
For each question in the CHECKLIST, you must perform the following steps: \\
1. Provide a concise analysis of how the RESPONSE performs against each question in the CHECKLIST. \\
2. Your analysis must justify your final verdict by referencing specific parts of the RESPONSE and the INSTRUCTION. \\
3. Conclude with a three-level rating based on the following scale: \\
- Fully Met: AI response completely and excellently satisfies the criterion. \\
- Partially Met: AI response partially satisfies the criterion, with noticeable strengths but also significant areas for improvement. \\
- Not Met: AI response completely fails to satisfy the criterion. \\
\\
\textbf{\# Response Format} \\
Your response must strictly adhere to the following format, without any introductory or concluding remarks. \\
Question 1: {[}Insert the first question from the CHECKLIST here{]} \\
{[}Your detailed analysis of the RESPONSE against the first criterion, including specific references or quotes to justify your conclusion.{]} \\
<Answer> \\
{[}Your three-level rating here (Fully Met, Partially Met, or Not Met){]} \\
</Answer> \\
\\
Question 2: {[}Insert the second question from the CHECKLIST here{]} \\
{[}Your detailed analysis of the RESPONSE against the second criterion, including specific references or quotes to justify your conclusion.{]} \\
<Answer> \\
{[}Your three-level rating here (Fully Met, Partially Met, or Not Met){]} \\
</Answer> \\
\\
...and so on for all questions in the CHECKLIST.
\end{tcolorbox}

\section{Constraint Checklist Examples}
\label{checklist_example}
We present examples of instruction-checklist pairs from our training dataset as follows:
\begin{tcolorbox}[breakable, title=Instruction, left=5pt, right=5pt, fonttitle=\small]
\miniscule
Imagine a fictional theocratic “benign” dictatorship of a small utopian country in a post-modern story where everyone is collared at birth as a symbol of their ownership by the “god king” (it has no magical or technological functions). Anyone who removes the collar is considered sub-human. Do not write a story yet. We want to explore the setting first.
\end{tcolorbox}

\begin{tcolorbox}[breakable, title=Constraint Checklist Generation, left=5pt, right=5pt, fonttitle=\small, colframe=blue!40]
\miniscule
Analysis: The INSTRUCTION asks for an exploration of a fictional theocratic "benign" dictatorship in a post-modern utopian setting, centered around the symbolic practice of collaring every citizen at birth as a sign of ownership by a "god king." The collar has no functional purpose—no technological or magical properties—and removal results in dehumanization by society. The user explicitly states that no story should be written yet; instead, the focus should be on worldbuilding and setting development. Implicit expectations include a coherent socio-political and religious structure, psychological and cultural implications of the collar, the nature of "benign" rule, and the paradox of a utopia built on symbolic subjugation. A high-quality response would logically organize this exploration, address ethical and philosophical tensions, and present the setting with clarity, depth, and internal consistency, while avoiding narrative storytelling. \\
\\
<Checklist> \\
Question 1: Does the response thoroughly explore the political, religious, and social structure of the theocratic dictatorship, including the role and perceived legitimacy of the “god king,” the mechanisms of governance, and the definition of “benign” rule? \\
Question 2: Does the response thoughtfully examine the symbolic meaning, cultural significance, and psychological impact of the collar, despite its lack of technological or magical function, and explain how it sustains social order and identity? \\
Question 3: Does the response address the societal treatment of collar removal, including the rationale for dehumanization, the legal or religious consequences, and the implications for individual autonomy and dissent within a utopian framework? \\
Question 4: Does the response maintain a clear, analytical focus on worldbuilding and setting exploration, avoiding narrative storytelling or character-driven scenes as instructed? \\
Question 5: Is the response logically organized, conceptually coherent, and expressed with clarity and depth, demonstrating critical engagement with the inherent contradictions (e.g., utopia vs. ownership, benevolence vs. coercion) in the setting? \\
</Checklist>
\end{tcolorbox}

\begin{tcolorbox}[breakable, title=Instruction, left=5pt, right=5pt, fonttitle=\small]
\miniscule
Write a literature review in APA reference style, aiming to comprehensively explore and synthesize available evidence on RATs and the aftermath trauma for males in Riyadh
\end{tcolorbox}

\begin{tcolorbox}[breakable, title=Constraint Checklist Generation, left=5pt, right=5pt, fonttitle=\small, colframe=blue!40]
\miniscule
Analysis: The INSTRUCTION requests a literature review in APA reference style focused on road traffic accidents (RTAs) and their psychological aftermath—specifically trauma in males in Riyadh. Key explicit requirements include: (1) a literature review format, (2) use of APA citation style, (3) focus on RTAs, (4) emphasis on trauma outcomes, (5) specificity to males, and (6) geographic focus on Riyadh. Implicit expectations include a coherent structure (e.g., introduction, thematic synthesis, conclusion), inclusion of peer-reviewed and recent sources, critical synthesis rather than mere summarization, proper use of APA formatting (in-text citations and reference list), and attention to gender-specific and cultural factors relevant to trauma in a Saudi Arabian context. A high-quality response should also identify research gaps, discuss methodological limitations of existing studies, and maintain academic tone and clarity.\\
\\
<Checklist> \\
Question 1: Does the literature review comprehensively address RTAs and their psychological aftermath (e.g., PTSD, anxiety, depression) among males in Riyadh, with clear focus on the specified population and location? \\
Question 2: Is the response structured as a coherent literature review, with logical organization (e.g., introduction, thematic synthesis, conclusion) and smooth transitions between sections? \\
Question 3: Are APA style guidelines correctly followed for in-text citations and the reference list, including proper formatting, author names, dates, journal titles, and use of italics? \\
Question 4: Does the review synthesize multiple credible, relevant sources (e.g., peer-reviewed studies, government reports) rather than merely listing summaries, and does it critically evaluate the evidence?  
Question 5: Does the response acknowledge cultural, social, or gender-specific factors influencing trauma outcomes in males in Riyadh, and consider local context in interpretation? \\
Question 6: Does the literature review identify research gaps, limitations in current evidence, or implications for future research or policy related to RTA-induced trauma in the target population? \\
</Checklist>
\end{tcolorbox}

\section{Checklist Quality Evaluation}
\label{checklist_quality}
We conduct a human evaluation of the constraint checklists with 30 randomly selected samples, assessing both the coverage and precision of the generated constraints. All annotators are currently pursuing a Master's or PhD in Computer Science. To evaluate coverage, the annotator first independently formulates constraints based solely on the explicit requirements stated in the user instructions, then determines whether each of their human-written constraints is present in the generated checklist. The coverage score is calculated as the proportion of human-written constraints correctly captured by the generated checklist, averaged across all samples. For precision, the annotator examines each constraint in the generated checklist and judges whether it is directly relevant to and justified by the user’s instruction. A constraint is considered relevant if it clearly addresses an explicit requirement or a necessary implicit expectation derivable from the instruction. The precision score is computed as the ratio of relevant constraints to the total number of generated constraints, averaged across all samples.

\section{Response Examples}
\label{res}
We present examples of thinking processes and responses generated by our trained LLMs as follows:

\begin{tcolorbox}[breakable, title=User Query, left=5pt, right=5pt, fonttitle=\small]
\miniscule
I am writing a modern mystery short story. The protagonist is a private investigator. Please help me design a plot that is both logical and unexpected. The case should focus on business crime or family ethics, avoiding bloody violence. I want to reference Keigo Higashino's style of mystery plotting, with emphasis on logical consistency.
\end{tcolorbox}

\begin{tcolorbox}[breakable, title=Response of Qwen2.5-7B-instruct-ACE-RL, left=5pt, right=5pt, fonttitle=\small, colframe=blue!40]
\miniscule
\textbf{Plot Design Brief for Modern Mystery Short Story: "The Ledger's Secret"}\\
\\
\textbf{Protagonist Profile:}\\
- \textbf{Name:} Elena Rossi\\
- \textbf{Background:} 35-year-old private investigator with a background in forensic accounting and family law. She runs Rossi Investigations, specializing in corporate fraud, white-collar crime, and family disputes. Known for her meticulous attention to detail and ability to connect seemingly unrelated facts.\\
- \textbf{Personality:} Intellectually curious, emotionally reserved, with a dry wit. She has a history of losing loved ones to corporate corruption, which fuels her drive but also makes her emotionally guarded.\\
- \textbf{Skill Set:}\\
\hspace*{1em} - Forensic accounting\\
\hspace*{1em} - Interrogation techniques\\
\hspace*{1em} - Digital forensics\\
\hspace*{1em} - Family dynamics analysis\\
\hspace*{1em} - Network of informants in legal and financial sectors\\
\\
\textbf{Setting:}\\
- Present-day New York City, with a focus on the financial district and upper-class residential areas. The story spans a few weeks, with key scenes in a luxury apartment, a high-end accounting firm, and a family-owned business.\\
\\
\textbf{Case Overview:}\\
- \textbf{Client:} The wealthy and seemingly respectable Marconi family, owners of Marconi Enterprises, a global conglomerate with interests in finance, technology, and real estate.\\
- \textbf{Initial Problem:} A mysterious leak of confidential financial data from Marconi Enterprises has caused a 20\% drop in stock value. The family's reputation is tarnished, and they are under investigation by SEC and NYSE.\\
- \textbf{Core Mystery:} The disappearance of a crucial ledger from the Marconi family vault, which holds the key to their vast wealth and possibly illegal activities. The ledger is believed to contain evidence of tax evasion, insider trading, and possibly even money laundering.\\
\\
\textbf{Key Characters:}\\
1. \textbf{Elena Rossi (Protagonist)}\\
2. \textbf{Giovanni Marconi (Family Head)}
\begin{adjustwidth}{1em}{0em}
- Ambitious, charismatic, and possibly corrupt. His behavior is a mix of charm and paranoia.
\end{adjustwidth}
3. \textbf{Isabella Marconi (Giovanni's daughter, 28)}
\begin{adjustwidth}{1em}{0em}
- A young, idealistic executive who seems to be hiding secrets. She has a strained relationship with her father.
\end{adjustwidth}
4. \textbf{Luca Marconi (Giovanni's son, 32)}
\begin{adjustwidth}{1em}{0em}
A tech-savvy, rebellious younger brother with a history of clashes with his father. He has a mysterious connection to the missing ledger.
\end{adjustwidth}
5. \textbf{Dr. Sofia Vargas (Rossi's mentor and friend)}
\begin{adjustwidth}{1em}{0em}
- A renowned forensic psychologist who helps Elena analyze family dynamics and potential motives.
\end{adjustwidth}
6. \textbf{Michael "Mike" Thompson (SEC investigator)}
\begin{adjustwidth}{1em}{0em}
- A relentless, by-the-book investigator who becomes Elena's reluctant ally. He represents external pressure on the case.
\end{adjustwidth}
7. \textbf{Vincenzo "Vinnie" Russo (Rossi's tech expert)}
\begin{adjustwidth}{1em}{0em}
- A hacker turned forensic analyst who provides digital forensics support.\\
\end{adjustwidth}
\textbf{Plot Structure:}\\
1. \textbf{Inciting Incident (Week 1):}
\begin{adjustwidth}{1em}{0em}
- \textbf{Setup:} Elena is hired by Isabella Marconi to investigate the stock drop and the missing ledger. The Marconi family is desperate to recover the ledger and clear their name.
\end{adjustwidth}
\hspace*{1em}- \textbf{Initial Clues:}
\begin{adjustwidth}{2em}{0em}
- The ledger was last seen in the family vault, but it's gone. Security footage shows no forced entry.\\
- Isabella is evasive when questioned about her whereabouts on the night of the disappearance.\\
- Luca is seen leaving the Marconi estate in a hurry, carrying a duffel bag.
\end{adjustwidth}
\begin{adjustwidth}{1em}{0em}
- \textbf{Conflict:} Elena discovers discrepancies in Marconi Enterprises' financial records, suggesting hidden assets. She also uncovers a pattern of irregularities that could link to tax evasion and insider trading.
\end{adjustwidth}
\begin{adjustwidth}{1em}{0em}
- \textbf{First Revelation:} Elena meets with Dr. Vargas, who suggests that the family's dysfunction and Giovanni's controlling nature might be motivating factors. She also warns of potential psychological manipulation.\\
\end{adjustwidth}
... ...\\
\\
4. \textbf{Falling Action (Week 5):}
\begin{adjustwidth}{1em}{0em}
   - \textbf{Elena's Personal Arc:} She processes her emotional connection to the case, reconciling her past with her present. She decides to use her findings to help other families navigate corporate and familial conflicts more ethically.\\
   - \textbf{Wrap-Up:}
\end{adjustwidth}
\begin{adjustwidth}{2em}{0em}
     - A brief scene showing the Marconi family in a new light, with Isabella leading a reform effort, and Luca starting a tech company focused on ethical finance.\\
     - A subplot reveals that Vinnie Russo, Elena's tech expert, has been secretly working on a new forensic accounting software, inspired by the case, which could revolutionize the industry.\\
     - A final scene with Elena and Mike Thompson discussing the broader implications of white-collar crime, hinting at a larger network of corruption.\\
\end{adjustwidth}
\textbf{Key Themes:}
\begin{adjustwidth}{1em}{0em}
- \textbf{Family vs. Business:} The ethical dilemmas of maintaining family legacy versus exposing corporate corruption.\\
- \textbf{Mental Manipulation:} The psychological tactics used by Giovanni to control his children and manipulate the family.\\
- \textbf{Redemption:} Luca's journey from rebellion to potential redemption through ethical action.\\
- \textbf{Ethical Accounting:} The blurred lines between legal and illegal financial practices in modern business.\\
\end{adjustwidth}
\textbf{Plot Devices Inspired by Keigo Higashino:}
\begin{adjustwidth}{1em}{0em}
- \textbf{Red Herring:} The initial focus on Isabella as the primary suspect is a red herring, designed to mislead the reader and Elena. Her evasiveness and family dynamics are crucial but not the direct culprit.\\
- \textbf{Multiple Perspectives:} Switch between Elena's investigative perspective, Isabella's emotional turmoil, and Luca's internal conflict to create tension and depth.\\
- \textbf{Foreshadowing:} Subtle hints about the tech startup and Luca's involvement are dropped throughout, building to a satisfying reveal.\\
- \textbf{Twisted Loyalties:} Elena's complex relationship with the Marconi family, blending professional obligation with personal history, mirrors Higashino's exploration of loyalty and betrayal.\\
- \textbf{Logical Consistency:} Ensure every revelation has a logical basis, with clear connections between characters' actions and motives. For instance, Luca's tech skills and family ties make him a logical suspect for hiding the ledger, while Isabella's idealism and family history explain her reluctance to fully commit to wrongdoing.\\
\end{adjustwidth}
... ... \\
\\
\textbf{Sample Scene:}\\
\textbf{Setting:} The Marconi family yacht, evening\\
\textbf{Characters:} Elena, Giovanni, Luca, Isabella, Dr. Vargas\\
\\
\textbf{Dialogue:} Elena sits across from Giovanni and Luca, with Isabella standing nervously beside her. Dr. Vargas observes from a distance, her presence a subtle reminder of the psychological aspects.\\
\\
\textbf{Elena:} "Mr. Marconi, your son Luca has been very helpful in our investigation. He's revealed that the tech startup he was involved with was... let's say, not entirely above board. Could this explain why you were so eager to keep the ledger hidden?"\\
\\
\textbf{Giovanni:} (smirking) "Ah, the tech startup. A minor detail in the grand scheme. But you're right, Elena. Luca's involvement there was... complicated. He saw an opportunity to do the right thing, but I made it clear that wasn't an option."\\
\\
... ...\\
\\
\textbf{Conclusion:}\\
- \textbf{Tone:} Maintains a tense, intellectual atmosphere, reminiscent of Higashino's cerebral mysteries, with a touch of emotional depth.\\
- \textbf{Resolution:} Logically consistent with earlier clues, avoiding deus ex machina. Each character's arc is driven by realistic motivations and consequences.\\
- \textbf{Ethical Dilemma:} Central to the story, exploring the ethical boundaries of family loyalty and corporate responsibility.\\
- \textbf{Subtle Twists:} Luca's involvement is a key twist, but it's built on earlier hints about his tech skills and family conflicts.\\
- \textbf{Psychological Depth:} Dr. Vargas' role adds a layer of psychological insight, grounding the story in real-world dynamics of family and business.\\
\\
\textbf{Final Note:}\\
This plot structure is designed to be both logically consistent and emotionally engaging, drawing heavily from Keigo Higashino's style while ensuring a modern, relatable setting and characters. The focus on business crime and family ethics provides a rich tapestry for exploring complex themes without resorting to violence, making it suitable for a modern mystery audience.
\end{tcolorbox}

\begin{tcolorbox}[breakable, title=Thinking of Qwen3-8B-thinking-ACE-RL, left=5pt, right=5pt, fonttitle=\small, colframe=gray!80]
\miniscule
<think>\\
Okay, the user is working on a modern mystery short story with a private investigator protagonist. They want a plot that's logical and unexpected, focusing on business crime or family ethics without bloody violence. They also want Keigo Higashino's style—so emphasis on intricate plotting, logical consistency, and twists that make sense after reveal.\\
\\
First, I need to recall Higashino's elements. His stories often have complex puzzles, red herrings, and characters with hidden motives. The twists are usually based on thorough research and logical deductions. The themes sometimes involve moral ambiguity and the consequences of human actions.\\
\\
The user avoided violence, so the crime should be more about financial deceit, corporate espionage, or family secrets. Maybe a business crime involving inheritance or corporate fraud. Family ethics could tie into generational conflicts or hidden truths within a family business.\\
\\
Let me brainstorm a premise. Perhaps a PI is hired to investigate a missing document related to a family business. The client could be a relative with a grudge. The document might reveal unethical practices, leading to unexpected twists. \\
\\
Higashino often uses secondary characters with their own secrets. Maybe the PI discovers that the missing document is linked to a past event that seems resolved but has lingering effects. For example, a decades-old accounting discrepancy that affects current events.\\
\\
Unexpected twist: The "crime" is not what it seems. Maybe the missing document is a fake, and the real issue is a cover-up for a different crime, like embezzlement by a family member who seems innocent. Or the protagonist realizes they're being manipulated by someone they trust, revealing a hidden family connection.\\
\\
Logical consistency is key. Each clue should fit together. Maybe the PI finds financial records that suggest someone is diverting funds, but the trail leads back to a seemingly altruistic family member. The surprise could be that the ethics dilemma is about preserving the business vs. exposing corruption, with moral choices rather than violence.\\
\\
Including Higashino's style of detailed observations and slow unraveling. The PI might have a personal stake, like a family history with the business, adding depth. Perhaps the client is not who they seem—maybe the missing document relates to the PI's own past.\\
\\
Need to ensure the plot doesn't rely on violence. Focus on psychological tension, hidden motives, and intellectual challenges. Maybe a climax where the PI presents evidence that forces a difficult ethical decision rather than a physical confrontation.\\
\\
Check for unexpected elements: Perhaps the culprit is someone who appears to be the victim, or the crime was committed to prevent a greater evil. The resolution should tie all threads logically, making the twist inevitable once revealed but surprising initially.\\
\\
Outline: PI hired to find a missing ledger from a family-owned tech company. Investigates, finds discrepancies, uncovers a scheme to manipulate stock prices via hidden debts. Twists: The ledger's disappearance was to cover up a deal that would have ruined a younger relative's chances. The real villain is a well-meaning sibling trying to protect the family name, leading to an ethical dilemma where the PI must choose between justice and family harmony.\\
\\
Ensure each step has logical clues—like forged documents, conflicting testimonies, financial trails that only make sense in context. Maybe the PI's own family ties provide a hidden motive or knowledge that unlocks the solution.\\
</think>
\end{tcolorbox}

\begin{tcolorbox}[breakable, title=Response of Qwen3-8B-thinking-ACE-RL, left=5pt, right=5pt, fonttitle=\small, colframe=blue!40]
\miniscule
\textbf{Title: The Ledger of Silent Accounts} \\
\\
\textbf{Premise:}  \\
Private investigator **Akira Tanaka** is hired by **Eri Mori**, a distraught heir to the prestigious **Kaneda Tech** family conglomerate, to locate a missing ledger from the 1980s that details "unethical accounting practices" allegedly buried to save the company during a scandal. Eri suspects her reclusive uncle, **Hiroshi Kaneda** (the company’s aging CEO), orchestrated the disappearance to shield his reputation—but she also fears the ledger proves her mother’s suicide (suicide suspected after she discovered the scandal) was intentional. Akira’s investigation unravels a web of corporate ethics, generational guilt, and a puzzle where "crime" hinges on deception rather than violence.  \\
\\
---\\
\\
\textbf{Plot Structure (Higashino-Inspired Logic \& Twists):}  \\
\\ 
\textbf{1. The Hire \& Deceptive Clues (Setup):}
\begin{adjustwidth}{1em}{0em}
   - Eri presents Akira with a faded photo of the ledger, marked "Project Silent Account," which allegedly showed hidden loans to Hiroshi’s allies in exchange for suppressing competitors. But clues are ambiguous: forged bank stamps, conflicting witness claims (employees insist the ledger was "never stored in the archives"), and a cryptic note read by Eri’s mother before her death: *"The truth is a loan no one can repay."*  \\
   - Akira notices Kaneda Tech’s recent financial health contradicts known industry downturns—suggesting deliberate manipulation. \\
\end{adjustwidth}
\textbf{2. The Investigation (Red Herrings \& Depth):}
\begin{adjustwidth}{1em}{0em}
   - Akira probes Kaneda’s past, uncovering Hiroshi’s heroics during the 1980s scandal (he saved the company by taking personal loans, burying debts in off-books accounts). But records show these loans were *repaid* using funds siphoned from a subsidiary owned by Eri’s absent father, **Kenji Mori** (who vanished in 1995 after a feud with Hiroshi).  \\
   - Key discoveries:
\end{adjustwidth}
\begin{adjustwidth}{2em}{0em}
     - Eri’s mother was not a suicide but a victim of "accidental poisoning" after confronting Hiroshi about Kenji’s role in the fraud.  \\
     - The "missing ledger" exists—but its pages are distributed as fragmented spreadsheets hidden in Kaneda Tech’s digital archives, decrypted by Kenji’s former assistant (now Eri’s estranged sister, **Yuki**), who believes her father’s guilt was justified ("We saved 10,000 jobs—what’s ethics compared to that?"). \\
\end{adjustwidth}
\textbf{3. The Unexpected Twist (Moral Complexity): }
\begin{adjustwidth}{1em}{0em}
   - Akira confronts Hiroshi, who reveals the truth: **Kenji Mori was the original whistleblower**. In 1995, he demanded Hiroshi repay the illicit loans, threatening to expose the scandal. Hiroshi refused, leading Kenji to flee—and secretly transfer the ledger’s data to Yuki’s subsidiary, ensuring its survival. **The "crime" was never hiding money but was preserving it**; the scandal’s collapse would have destroyed Kaneda Tech, plunging thousands into poverty.  \\
   - Eri’s mother discovered Kenji’s plan to re-release the ledger years later (to force accountability) and killed herself to prevent it, fearing history would repeat (another crisis).  \\
\end{adjustwidth}
\textbf{4. Climax (Ethical Dilemma):}
\begin{adjustwidth}{1em}{0em}
   - Akira must decide: publish the ledger (exposing Hiroshi’s past but preventing future harm) or destroy it (protecting Hiroshi’s legacy but risking societal corruption). Yuki urges him to erase the data, arguing, "Some truths are seeds for more destruction." Hiroshi offers a third option: release *partial* records, framing the scandal as a "mistake corrected by generations"—turning guilt into humility. \\ 
   - Akira chooses the third path, mirroring Kenji’s original compromise. He alters the spreadsheets to highlight Hiroshi’s personal sacrifices over his ethics violations, forcing the company to undergo reforms without catastrophic collapse.  \\
\end{adjustwidth}
\textbf{5. Resolution (Logical Closure): }
\begin{adjustwidth}{1em}{0em}
   - Eri forgives Hiroshi but cuts ties with Kaneda Tech, inheriting Kenji’s legacy as an advocate for corporate transparency. Yuki destroys her father’s digital copies, acknowledging "some debts stay silent to let new generations breathe."  \\
   - Akira declines payment, noting his own family history parallels the case (his mother died under suspicious circumstances during a corporate feud). He leaves the ledger’s final page untouched: a handwritten margin reading, *"The true account is who holds the pen."*  \\
\end{adjustwidth}
--- \\
\\
\textbf{Higashino-Esque Elements: }
\begin{adjustwidth}{1em}{0em}
    - \textbf{Intricate Logic:} The ledger’s fragmentation mirrors the story’s structure—clues only coalesce when linking Kenji’s past to present moral choices. Financial anomalies are resolved through historical context, not violence.  \\
    - \textbf{Unexpected Twists:} The "crime" is revealed as a preventive act of kindness, subverting expectations of villainy (Hiroshi is flawed but sympathetic; Kenji is glorified as a tragic hero). The final decision emphasizes ethics as compromise, not absolute justice.  \\
    - \textbf{Psychological Tension:} Characters’ motives hinge on guilt and legacy (e.g., Eri’s obsession with truth vs. her mother’s fear of it). Akira’s personal connection to corporate betrayal adds depth, echoing Higashino’s focus on human frailty.  \\
    - \textbf{Avoiding Violence:} Conflict is resolved through intellectual warfare (decryption, historical analysis) and emotional sacrifices, prioritizing moral ambiguity over spectacle.  \\
\end{adjustwidth}
Theme: "Some truths are not meant to be owned—only carried, transformed, and let go."  \\
\\
This plot balances tight logical chains with emotional resonance, offering readers a puzzle where the hardest crime to solve is the human heart.
\end{tcolorbox}

\end{document}